\documentclass[final,5p,times,twocolumn,authoryear]{elsarticle}

\usepackage{amssymb}
\usepackage{algorithm}
\usepackage{algpseudocode}
\usepackage{graphicx}

\usepackage{subfigure}
\usepackage{amsmath}
\usepackage{url}

\usepackage{amssymb}
\usepackage{xcolor}
\usepackage{tabularx} 
\usepackage{multirow}
\usepackage{tabu}
\usepackage{pbox}
\usepackage{amssymb}
\usepackage{amsfonts}
\usepackage{bbm}
\usepackage{graphicx}
{\begin{list}               
    {$\bullet$ \hfill}{
        \setlength{\leftmargin}{\parindent}
        \setlength{\parsep}{0.04\baselineskip}
        \setlength{\itemsep}{0.5\parsep}
        \setlength{\labelwidth}{\leftmargin}
        \setlength{\labelsep}{0em}}
    }
{\end{list}}

\providecommand{\eref}[1]{Eq. \eqref{#1}}  
\providecommand{\cref}[1]{Chapter~\ref{#1}}
\providecommand{\sref}[1]{Section~\ref{#1}}
\providecommand{\fref}[1]{Figure~\ref{#1}}
\providecommand{\tref}[1]{Table~\ref{#1}}

\renewcommand{\vec}[1]{\ensuremath{\boldsymbol{#1}}}
\providecommand{\mat}[1]{\ensuremath{\boldsymbol{#1}}}

\providecommand{\calL}{\mathcal{L}}

\providecommand{\mF}{\mat{F}}

\providecommand{\mI}{\mat{I}}

\providecommand{\vp}{\vec{p}}

\providecommand{\vy}{\vec{y}}

\journal{Engineering Applications of Artificial Intelligence}

\begin{document}

\begin{frontmatter}

\title{Improving Weakly-Supervised Object Localization Using Adversarial Erasing and Pseudo Label}

\author[1]{Byeongkeun Kang}
\ead{byeongkeun.kang@seoultech.ac.kr}

\author[1]{Sinhae Cha}
\ead{csh27@seoultech.ac.kr}

\author[2]{Yeejin Lee\corref{cor1}}
\ead{yeejinlee@seoultech.ac.kr}
\cortext[cor1]{Corresponding author.}
\affiliation[1]{organization={Department of Electronic Engineering, Seoul National University of Science and Technology},
            addressline={232 Gongneung-ro, Nowon-gu}, 
            city={Seoul},
            postcode={01811}, 
            country={South Korea}}
\affiliation[2]{organization={Department of Electrical and Information Engineering, Seoul National University of Science and Technology},
            addressline={232 Gongneung-ro, Nowon-gu}, 
            city={Seoul},
            postcode={01811}, 
            country={South Korea}}

\begin{abstract}
Weakly-supervised learning approaches have gained significant attention due to their ability to reduce the effort required for human annotations in training neural networks. This paper investigates a framework for weakly-supervised object localization, which aims to train a neural network capable of predicting both the object class and its location using only images and their image-level class labels. The proposed framework consists of a shared feature extractor, a classifier, and a localizer. The localizer predicts pixel-level class probabilities, while the classifier predicts the object class at the image level. Since image-level class labels are insufficient for training the localizer, weakly-supervised object localization methods often encounter challenges in accurately localizing the entire object region. To address this issue, the proposed method incorporates adversarial erasing and pseudo labels to improve localization accuracy. Specifically, novel losses are designed to utilize adversarially erased foreground features and adversarially erased feature maps, reducing dependence on the most discriminative region. Additionally, the proposed method employs pseudo labels to suppress activation values in the background while increasing them in the foreground. The proposed method is applied to two backbone networks (MobileNetV1 and InceptionV3) and is evaluated on three publicly available datasets (ILSVRC-2012, CUB-200-2011, and PASCAL VOC 2012). The experimental results demonstrate that the proposed method outperforms previous state-of-the-art methods across all evaluated metrics.
\end{abstract}


\end{frontmatter}



\section{INTRODUCTION} \label{sec:introduction}
Object localization is an essential task in various computer vision applications such as robot grasping, navigation, and intelligent systems~\citep{Cai2023_EAAI, Afaq2022_r3, Li2023_EAAI}. Since it involves localizing and classifying an object in an input image, it is used to localize a target object for robot grasping. In robot navigation, it is utilized to localize road participants and to reach a target location safely.

\begin{figure}[!t] \begin{center}
\begin{minipage}{0.49\linewidth}
\centerline{\includegraphics[scale=0.53]{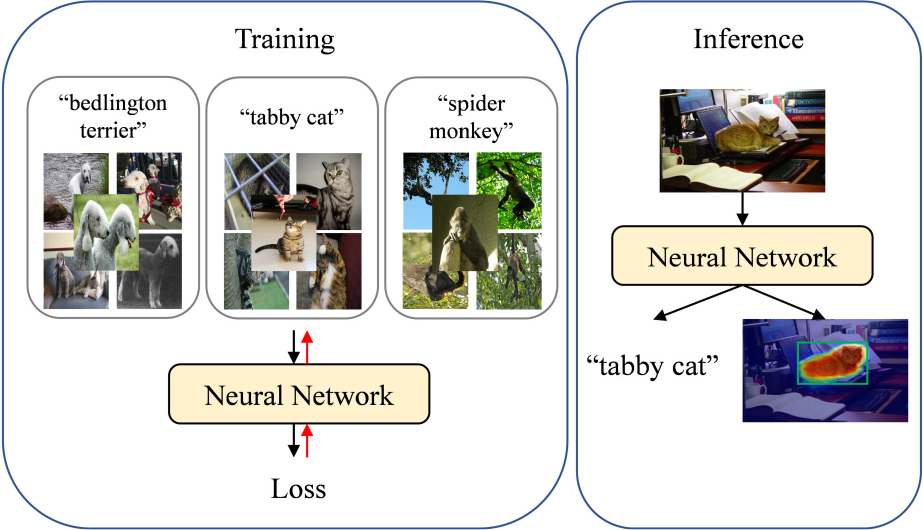}}
\end{minipage}
   \caption{Illustration of weakly-supervised object localization.}
\label{fig:teaser_task}
\end{center}\end{figure}

However, training an object localization method using supervised learning requires a training dataset consisting of not only images but also a class label and a bounding box label per image. Since class and bounding box labels are typically annotated by humans, they are expensive and limit the diversity of classes in object localization datasets. To overcome this limitation, researchers have investigated weakly-supervised object localization methods that only require images and image-level class labels~\citep{Wu_2022_CVPR}, as described in~\fref{fig:teaser_task}. Because annotating image-level class labels demands far less effort compared to annotating bounding box labels, these methods significantly reduce the annotation efforts and enable localizing diverse object classes.



Given the importance of reducing annotation efforts~\citep{jang2023Weakly, Liu2023_EAAI}, many researchers have explored various approaches for weakly-supervised object localization. \cite{Oquab2015} introduced one of the earliest methods for utilizing convolutional neural networks (CNNs) in weakly-supervised object localization. They investigated the possibility of achieving accurate object localization without the need for bounding box labels. While \cite{Oquab2015} localized a point for each object using max pooling, \cite{cam2016} proposed a method that predicts the full extent of objects by employing average pooling instead of max pooling. They introduced the concept of a class activation map (CAM), which can be interpreted as the probability of each class at each pixel. CAM is generated by the weighted accumulation of activation maps from CNNs. Although this method demonstrated the potential for learning object localization without bounding box labels, it typically localizes only the most discriminative part of objects due to the lack of detailed location information during training.

Therefore, researchers have explored advanced methods to localize the entire object region rather than only the most discriminative region. These methods include utilizing multiple feature maps~\citep{Xue2019}, employing adversarial erasing~\citep{ACoL2018}, incorporating pseudo labels~\citep{Zhang2020}, and designing alternative architectures~\citep{Gupta_2022_CVPR}. Adversarial erasing has been employed to reduce reliance on the most discriminative region of an object by erasing the corresponding content in an image. 

\begin{figure}[!t] \begin{center}
\begin{minipage}{0.49\linewidth}
\centerline{\includegraphics[scale=0.53]{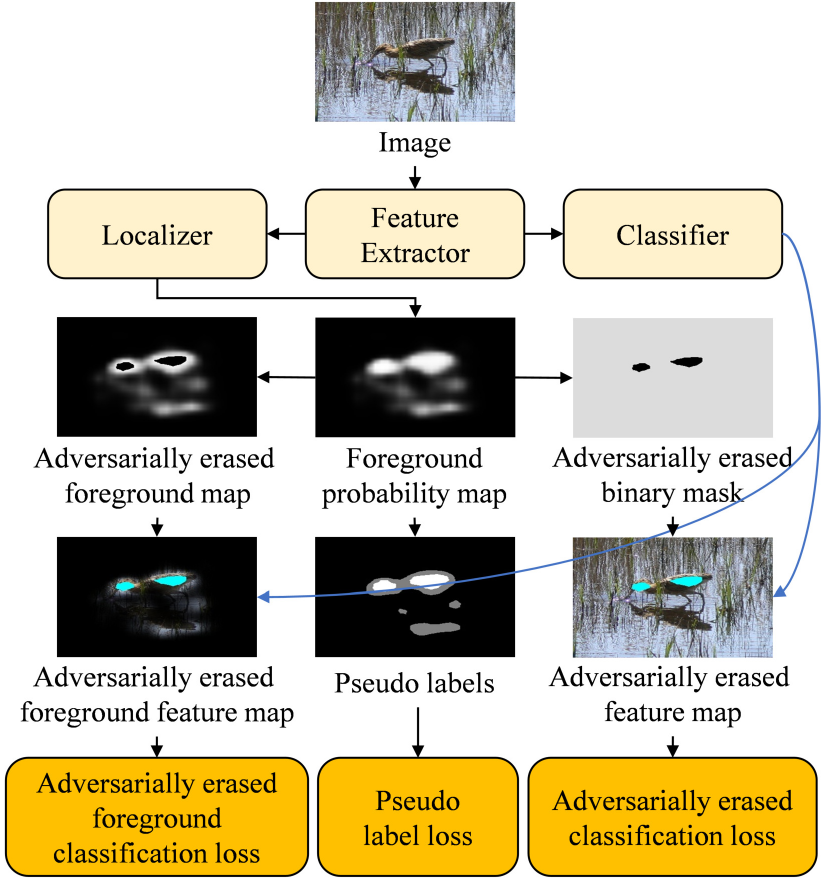}}
\end{minipage}
   \caption{Illustration of the proposed two adversarial erasing-based losses and a pseudo-label-based loss. While we apply adversarial erasing to feature maps from the classifier rather than the image itself, the erased regions are overlaid on the image and denoted by the cyan color for better visibility. The loss on the rightmost side is computed using the feature map where discriminative regions are erased. The loss on the leftmost side takes features only from the foreground region with erasing to train the localizer. Pixel-level pseudo labels are generated and utilized as an auxiliary loss to suppress the background while activating the entire object region.}
\label{fig:teaser2}
\end{center}\end{figure}

While recent works have explored methods to enhance localization by introducing additional processing~\citep{ACoL2018} or utilizing separate networks such as a bounding box regressor~\citep{Guo2021} or a pseudo-label generator~\citep{Zhang2020}, we focus on improving localization performance without adding extra processing or network branches during inference. The proposed method centers on introducing novel and enhanced loss terms for training, which include an improved implicit adversarial erasing method and a pseudo-label generation approach, as shown in~\fref{fig:teaser2}.

In more detail, whereas \cite{ACoL2018} explicitly erased the most discriminative region and trained a separate network, our approach implicitly encourages the network to reduce its dependency on the discriminative region and localize the entire object region, all without using a separate network. Additionally, we introduce two complementary erasing methods, one applied to the score map for classification and the other to the foreground mask. Experimental results demonstrate that they are complementary. While \cite{Zhang2020} generated pseudo bounding box labels and trained a class-agnostic regressor, we generate pixel-level pseudo activation maps and utilize them to calculate a loss. In contrast to the method presented in~\citep{spg2018}, we utilize the activation map for the ground-truth class to compute the loss rather than multiple intermediate feature maps.

As weakly-supervised object localization methods are trained solely using images and their corresponding image-level class labels, the required training data is identical to that of an image classification task. Consequently, these methods can be trained using any image classification dataset. We also leverage this advantage following previous literature~\citep{Wu_2022_CVPR} and use three image classification datasets to demonstrate the performance of our proposed framework.

\begin{figure}[!t] \begin{center}
\begin{minipage}{0.49\linewidth}
\centerline{\includegraphics[scale=0.65]{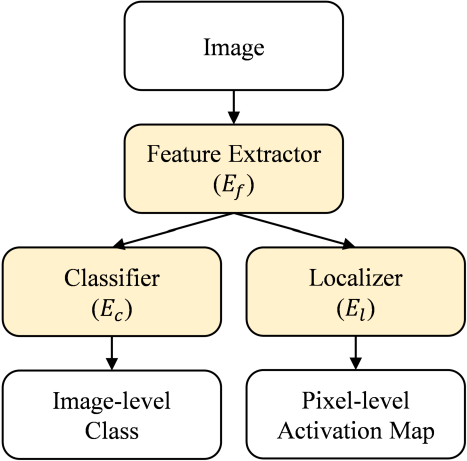}}
\end{minipage}
   \caption{Illustration of the proposed framework. Given an input image, it aims to predict an image-level class label and a pixel-level activation map. The pixel-level activation map can be interpreted as the probability of the existence of a specific object at each pixel. Given an input image, a feature extractor first processes it to encode a shared feature representation for both classification and localization. Then, the feature map is processed by a classifier and a localizer, separately.}
\label{fig:overview}
\end{center}\end{figure}

In this paper, we introduce a weakly-supervised object localization (WSOL) framework that consists of three main modules, a feature extractor, a classifier, and a localizer, as illustrated in~\fref{fig:overview}. We then train the network using a novel loss function that is based on adversarial erasing, pseudo-label generation, and foreground prediction. The effectiveness of our proposed framework is demonstrated through experiments using three publicly available datasets (ILSVRC-2012~\citep{ILSVRC15}, CUB-200-2011~\citep{cub_dataset}, and PASCAL VOC 2012~\citep{pascalvoc}).

The contributions of this paper are as follow: 
\begin{itemize}
  \item We propose a weakly supervised object localization framework that is trained using a novel loss function and achieves state-of-the-art performance.
  \item The proposed loss function is based on adversarial erasing, pseudo-label generation, and foreground prediction. Specifically, it is calculated by effectively using adversarially erased feature maps, adversarially erased foreground masks, pseudo labels, foreground masks, and image-level class labels.
  \item We demonstrate the superiority of the proposed method over the previous state-of-the-art approach using three standard metrics (Top-1, Top-5, and GT-known) and an additional metric, mean Intersection over Union (mIoU). We employ the mIoU metric to provide a more detailed assessment of localization performance. These evaluations are conducted across two backbone networks and three public datasets.
\end{itemize}


\section{RELATED WORKS}
To train object localization methods without bounding box labels, researchers proposed weakly supervised object localization methods that are trained using only images and image-level class labels. These methods learn to predict a corresponding object class given an image similar to an image classification task. The difference is that they are additionally expected to find the location of the object in contrast to those for image classification. 

\cite{Oquab2015} introduced one of the earliest works on weakly-supervised object localization. They proposed a method that can localize an object per class and per image by training a CNN using only image-level class labels. The method utilized a sliding window and a global max-pooling to search the locations of objects. They evaluated the approach by comparing the location of the maximum score to ground-truth bounding box labels. \cite{cam2016} proposed to utilize a global average pooling layer instead of a global max-pooling layer. After training a CNN with a global average pooling layer using a dataset for an image classification task, their method generates a class activation map (CAM) by a linear combination of feature maps. As CAM indicates discriminative regions for each object class, they utilized CAM to estimate the location of each object. They computed a localization error by comparing predicted bounding boxes to ground-truth bounding box labels. \cite{gradcam2017} proposed the Grad-CAM method that can be applied to any neural network to obtain CAM rather than typical image classification networks.

As typical CAM usually indicates only the most discriminative part, researchers proposed methods to reduce dependence on the regions by erasing them. \cite{hideseek2017} presented an approach to include other relevant parts in localization. They proposed to hide partial regions in training images randomly so that a network cannot depend on only the most discriminative part for classification. To explicitly reduce dependence on highly distinct regions, \cite{ACoL2018} proposed to utilize two heads in a network. One is a typical classifier that takes feature maps from a backbone in the network. The other takes erased feature maps where discriminative regions are found and are erased by the former classifier. \cite{Choe2019} proposed the attention-based dropout layer (ADL) which utilizes either a drop mask or an importance map~\citep{Choe2021}. When a drop mask is applied, it hides the most discriminative region similar to the ACoL~\citep{ACoL2018}. The importance map is used to depend more on informative regions similar to an attention mechanism~\citep{cbam2018}. 

To localize whole region of objects rather than partial region, researchers also investigated approaches that effectively utilize multiple feature maps together. \cite{Xue2019} presented the DANet that utilizes hierarchical divergent activation (HDA) and discrepant divergent activation (DDA). The HDA uses hierarchical classes to activate not only the most discriminative part but also other shared features between classes. The DDA is to force multiple activation maps to discover varying components of objects. \cite{WU202223_r2} extended this method for weakly-supervised object detection. \cite{Yang2020} proposed to utilize activation maps from all the classes together rather than relying on only the map of the class with the highest probability, to suppress background better. They also utilized low- and high-level feature maps together to avoid localizing only the most discriminative region. \cite{Jiang2021} also presented a method that generates activation maps at varying layers and fuses them to obtain a more reliable map. \cite{Gao2023_r2} proposed the DB-HybridNet that also fuses high- and low-level features and concatenates the combined features. The network then employs the broad learning system (BLS)~\citep{Chen2018_BLS} to produce optimal outputs from the concatenated features. \cite{Wu_2022_CVPR} proposed the background activation suppression (BAS) method that suppresses background activation values to learn the whole region of objects. They also utilized activation maps of other classes to determine final localization result in an inference stage similar to the NL-CCAM~\citep{Yang2020}. \cite{Wei2021} proposed to utilize element-wise multiplication of shallow and deep features to obtain sharp boundaries. They also proposed to use class-agnostic segmentation model that is trained by using pseudo labels from CAM.

\begin{table*}[!t]
\centering
\begin{minipage}{0.99\linewidth}
\caption{Summary of relevant and selected previous works in weakly-supervised object localization.}
\label{tab:related_works}
\renewcommand{\arraystretch}{1.} 
\centering
\small
\begin{tabular}{ >{\centering}m{0.23\textwidth}| >{\centering}m{0.10\textwidth}| >{\raggedright\arraybackslash}m{0.57\textwidth} } 
\hline
Method & Category & \multicolumn{1}{c}{Summary}   \\ 
\hline\hline
HaS~\citep{hideseek2017} & & - Random erasing applied to images \\
\cline{1-1} \cline{3-3} 
 & & - Adversarial erasing applied to feature maps \\
ACoL~\citep{ACoL2018} & & - Two parallel branches and additional processing in inference \\
 & \multirow{2}{*}{Erasing} & \hspace{0.15cm} (one branch for the entire feature map, the other for erased feature map) \\ 
\cline{1-1} \cline{3-3} 
ADL~\citep{Choe2019} & & - Random selection between attention mechanism and adversarial erasing \\
\cline{1-1} \cline{3-3} 
 &  & - Two adversarial erasing-based losses (one using erased feature map, \\ 
Proposed &  & \hspace{0.15cm} the other using erased foreground feature map) \\ 
 &  & - No additional processing/branch in inference   \\ 
\hline  	
\multirow{2}{*}{SPG~\citep{spg2018}} & & - Fusion of multiple branches at varying levels for localization, \\
		&  & \hspace{0.15cm} trained using pixel-level binary cross-entropy losses \\
\cline{1-1} \cline{3-3} 
PSOL~\citep{Zhang2020} &  & - Class-agnostic bounding box regressor trained using pseudo box labels  \\
\cline{1-1} \cline{3-3} 
SLT~\citep{Guo2021} & Pseudo label & - Bounding box regressor trained using pseudo box labels \\
\cline{1-1} \cline{3-3} 
 &  & - One pseudo-label-based loss using pixel-level $\ell_1$ distance \\ 
Proposed &  & \hspace{0.15cm} (pseudo labels are generated using only foreground probability map) \\ 
 &  & - No separate bounding box regressor or separate pseudo label generator   \\ 
\hline  	
\multirow{2}{*}{SLT~\citep{Guo2021}}  &  & - Entirely separated classifier and bounding box regressor \\ 
			 &  & - Loss enforcing consistency after forward and inverse transforms \\
\cline{1-1} \cline{3-3} 
\multirow{2}{*}{SPA~\citep{Pan2021}} & Others & - Loss using activation values to suppress background  \\
 &  & - Post-processing using self-correlation  \\
\cline{1-1} \cline{3-3} 
BAS~\citep{Wu_2022_CVPR} &  & - Loss using activation values in addition to cross-entropy losses \\
\hline
\end{tabular}
\end{minipage}
\end{table*}

Pseudo labels were also generated and utilized to improve the performance of localization. \cite{spg2018} proposed to train a CNN by utilizing self-produced guidance (SPG) masks. They generated SPG masks by considering pixels with high/low scores as foreground/background and by ignoring pixels with intermediate scores. \cite{Zhang2020} also proposed a pseudo supervised object localization method. They generated pseudo bounding box labels by using either weakly-supervised object localization methods or deep descriptor transforming (DDT)~\citep{DDT2019}. They then trained a class-agnostic regressor using the pseudo bounding box labels for localization. \cite{Guo2021} proposed to separate localization and classification networks and to utilize pseudo ground-truth for training the localization network similar to~\citep{Zhang2020}. It also utilizes data augmentation methods and their inverse transforms. Specifically, it forces that the activation map from an input image matches the inverse transformed activation map of the transformed image. \cite{Pan2021} proposed the restricted activation module (RAM) to constrain the range of values of activation to avoid high dependence on local extreme activation values. RAM also comprises utilizing a pseudo mask to discriminate the foreground from the background better. They also proposed a post-processing method that employs self-correlation between feature vectors. As mentioned earlier, \cite{Wei2021} also utilized pseudo labels.

Methods employing vision transformers rather than typical CNNs were also proposed~\citep{Kim2023Multiscale}. \cite{Gao2021} proposed a transformer~\citep{visionTransformer2020}-based method to utilize long-range feature dependency. \cite{Gupta_2022_CVPR} also investigated a method that utilizes the vision transformer and patch-based dropout layer.

Other novel approaches were also proposed such as formulating the weakly-supervised object localization problem into a contrastive learning problem or a domain adaptation problem. \cite{i2c2020} presented the inter-image communication (I2C) method. Features from objects that belong to the same class are forced to be similar within each mini-batch and also across an entire dataset. \cite{Kim2021} proposed a method that explicitly utilizes a latent variable for location. It hence can visualize various attribution maps. \cite{Zhu_2022_CVPR} approached the weakly supervised object localization task as a domain adaptation problem. They considered images as a source domain and pixels/patches as a target domain. \cite{Xie2021} proposed a two-stage framework to train a network for localization. The first stage trains the network to generate a coarse activation map from low-level features. Then, the second stage fine-tunes the network to generate a better-quality activation map. \cite{Xie_2022_CVPR} proposed to utilize a class-agnostic foreground-background separation map. They trained a network to separate foreground regions from the background by using contrastive learning. Then, the separation map is utilized to suppress the background area in a class activation map. \cite{normalization2021} investigated methods for normalizing CAM to achieve higher accuracy. Recently, \cite{fdcnet} presented a method that incrementally learns object localization using only image-level class labels. \tref{tab:related_works} summarizes the relevant previous works in weakly-supervised object localization. For a more comprehensive survey, we refer readers to~\citep{Zhang2022_r2}.

The proposed method utilizes adversarial erasing to reduce dependency on highly discriminative regions similar to~\citep{ACoL2018, Choe2019, Choe2021}. However, while the method in~\citep{ACoL2018} explicitly generates two activation maps, one for the highly discriminative part and the other for the less discriminative region, the proposed method implicitly guides the network to activate an entire object. Accordingly, the method~\citep{ACoL2018} needs to fuse the maps while the proposed method does not need. Furthermore, we propose two different erasing methods that are erasing the score map for classification and erasing the foreground probability mask. We experimentally demonstrate that the two methods are complementary in~\sref{sec:results}.  

The proposed method also computes a loss term using pseudo labels. The loss is computed by using $\ell_1$ distance rather than a binary cross-entropy loss function in~\citep{spg2018, Wei2021}. Moreover, the proposed method calculates the loss by using the activation map for the ground-truth class while \cite{Wei2021} and \cite{spg2018} use a class-agnostic foreground mask and multiple intermediate feature maps, respectively.

\section{PROPOSED METHOD}
\label{sec:method}
The goal of the proposed method is to train a neural network that can predict both object class and location by using only image-level class label. As only images and their image-level class labels are given, the provided data is the same as that for an image classification task. Also, since the networks for an image classification task are studied well, we utilize them for the backbone of the proposed method. Given a backbone, the head of the network is replaced by one for classification and the other for localization following the BAS~\citep{Wu_2022_CVPR}. The details of the network architecture are explained in~\sref{sec:wsol_network}. Then, to train the network for both classification and localization, novel loss terms are proposed in~\sref{sec:loss}. 

\begin{figure}[!t] 
\begin{center}
\begin{minipage}{0.49\linewidth}
\centerline{\includegraphics[scale=0.52]{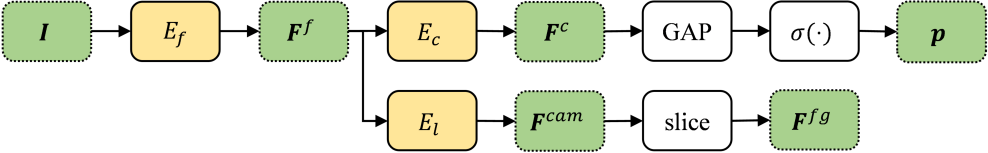}}
\end{minipage}
   \caption{The proposed framework during inference. $E_f$, $E_c$, and $E_l$ represent a feature extractor, a classifier, and a localizer, respectively. $\mI$, $\mF^{f}$, $\mF^{c}$, $\mF^{cam}$, and $\mF^{fg}$ denote an input image, a feature map, a score map, and a class activation map, and a foreground mask, respectively. $\vp$ represents a probability vector for classification. GAP, $\sigma(\cdot)$, and slice denote a global average pooling layer, a softmax function, and extracting a channel from a tensor, respectively.}
\label{fig:framework_inference}
\end{center}
\end{figure}

\subsection{WSOL Network}
\label{sec:wsol_network}
The network consists of three major modules, a feature extractor $E_f$, a classifier $E_c$, and a localizer $E_l$ as shown in~\fref{fig:framework_inference}. Given an image $\mI$, it is first processed by the feature extractor $E_f$ to encode a meaningful feature representation $\mF^f \in \mathbb{R}^{28 \times 28 \times 512}$ for both classification and localization. To achieve accurate localization, the feature representation $\mF^f$ is extracted to have a higher resolution compared to those from typical architectures for classification since maintaining spatial information is more important than higher-level information for localization. It is achieved by re-organizing the layers between the last pooling layer of the $E_f$ and the final output of the $E_c$.

The feature map $\mF^f$ is then fed into the classifier $E_c$ and the localizer $E_l$, separately. 
The classifier $E_c$ processes $\mF^f$ and outputs a score map $\mF^c \in \mathbb{R}^{14 \times 14 \times C}$ where $C$ denotes the total number of classes. The map $\mF^c$ is then processed by a global average pooling to reduce the spatial dimension and by a softmax function to constrain the range of the values. Finally, the output of the softmax function estimates the probability $\vp \in \mathbb{R}^{C}$ of being each class. 

The localizer $E_l$ also takes $\mF^f$ and then outputs the class activation map $\mF^{cam} \in \mathbb{R}^{28 \times 28 \times C}$. 
Given the class activation map $\mF^{cam}$, the foreground mask $\mF^{fg} \in \mathbb{R}^{28 \times 28}$ is obtained by extracting a channel of the map $\mF^{cam}$ where the channel corresponds to either the ground-truth class or a predicted class. In an inference stage, the foreground mask $\mF^{fg}$ is interpolated to the resolution of the image $\mI$ to represent the pixel-level score map for containing an object of a specific class.

\subsection{Training WSOL Network} \label{sec:loss}
To achieve high accuracy by training the proposed network, we propose a novel loss function that is based on adversarial erasing, pseudo-label generation, and foreground prediction (see~\fref{fig:framework_training}). The loss function consists of seven loss terms including one typical image classification loss, two using adversarial erasing, one using pseudo labels, and others. The effectiveness of the proposed loss terms are demonstrated in~\sref{sec:results}. Details of each loss term are explained as follows:

\begin{figure*}[!t] 
\begin{center}
\begin{minipage}{0.49\linewidth}
\centerline{\includegraphics[scale=0.60]{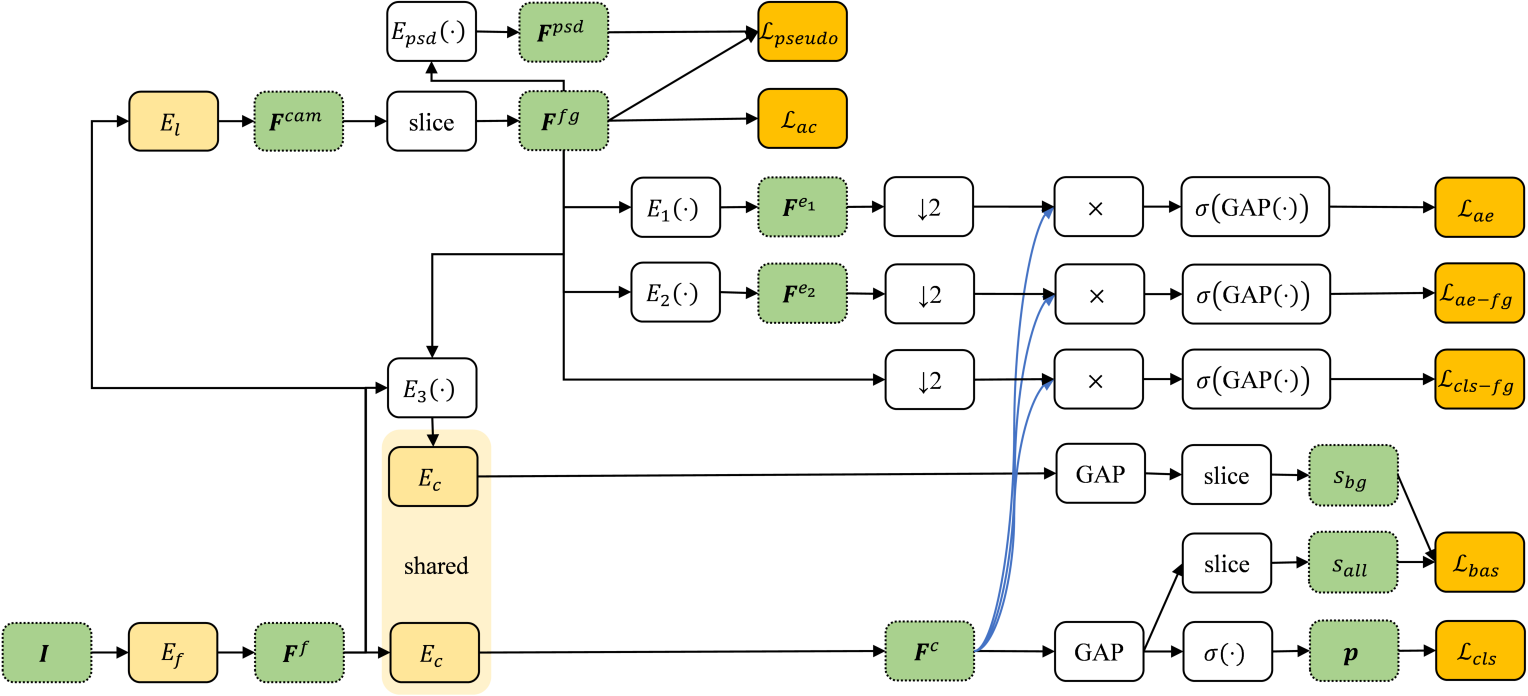}}
\end{minipage}
   \caption{The proposed framework during training. $E_f$, $E_c$, and $E_l$ represent a feature extractor, a classifier, and a localizer, respectively. $\mI$, $\mF^{f}$, $\mF^{c}$, $\mF^{cam}$, and $\mF^{fg}$ denote an input image, a feature map, a score map, and a class activation map, and a foreground mask, respectively. $\vp$ represents a probability vector for classification. GAP, $\sigma(\cdot)$, and slice denote a global average pooling layer, a softmax function, and extracting a channel from a tensor, respectively.}
\label{fig:framework_training}
\end{center}
\end{figure*}

\noindent \textbf{Classification loss $\calL_{cls}$.} It is a typical cross-entropy loss that is used in an image classification task. It compares a predicted probability vector $\vp$ for classification to the corresponding image-level class label $\vy$. 
The classification loss $\calL_{cls}$ is computed as follows:
\begin{equation}
	\calL_{cls} = - \sum_{i=1}^{C} \vy_i  \ln({ \sigma(\text{GAP}( \mF^c ))_{i}}) = - \sum_{i=1}^{C} \vy_i  \ln{\vp_{i}}
\label{eq:loss_class}
\end{equation}
where $C$ denotes the total number of classes in a dataset; GAP and $\sigma(\cdot)$ represent a global average pooling and a softmax function, respectively.

\noindent \textbf{Foreground classification loss $\calL_{cls\text{-}fg}$.}
It is to guide the localizer as well as the classifier by using a background suppressed score map $\hat{\mF}^{c}$. 
To compute a background suppressed score map, the score map $\mF^c$ from the classifier and the predicted foreground mask $\mF^{fg}$ are utilized. 
As the spatial dimension of $\mF^{fg}$ is higher than that of $\mF^c$, the foreground mask $\mF^{fg}$ is first downsampled by a factor of two by using an average pooling. Then, the background suppressed score map $(\hat{\mF}^{c} \in \mathbb{R}^{14 \times 14 \times  C})$ is obtained by concatenating the element-wise multiplication of each channel of $\mF^c$ and the downsampled foreground mask $\hat{\mF}^{fg}$. It is then processed by a global average pooling layer and a softmax function similar to~\eref{eq:loss_class}, to obtain the probability vector $(\hat{\vp} \in \mathbb{R}^{C})$ from the foreground region.
Then, the foreground classification loss $\calL_{cls\text{-}fg}$ is computed by comparing it to the image-level class label $\vy$. 
\begin{equation}
\begin{split}
	\calL_{cls\text{-}fg} & = - \sum_{i=1}^{C} \vy_i  \ln \big(\sigma(\text{GAP}( \mF^c \cdot \hat{\mF}^{fg} ))_{i} \big) \\
									   & = - \sum_{i=1}^{C} \vy_i  \ln{\hat{\vp}_{i}} 
\end{split}
\label{eq:loss_foreground}
\end{equation}

\noindent \textbf{Adversarially erased classification loss $\calL_{ae}$.}
While~\cite{ACoL2018} explicitly trained a separate branch using adversarial erasing and used it in inference to localize the entire object region, we implicitly guide a single localizer to localize the entire region using two adversarial erasing-based loss terms. Therefore, we do not require any additional processing or branch in inference.

Both adversarial erasing-based loss terms aim to reduce the dependency on the most discriminative region and increase activation values on the less discriminative region of the object. Since cross-entropy losses can approach zero by solely relying on the most discriminative region, we erase the corresponding region in the feature map so that the less discriminative regions have a larger influence in the loss computation.

\begin{figure}[!t] 
\begin{center}
\begin{minipage}{0.32\linewidth}
\centerline{\includegraphics[scale=0.30]{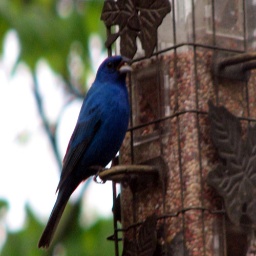}}
\end{minipage}
\begin{minipage}{0.32\linewidth}
\centerline{\includegraphics[scale=0.30]{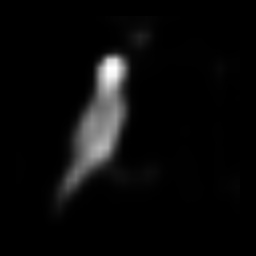}}
\end{minipage}
\begin{minipage}{0.32\linewidth}
\centerline{\includegraphics[scale=0.30]{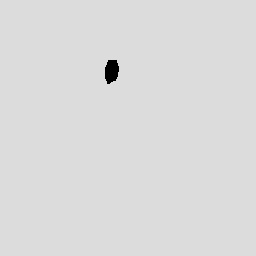}}
\end{minipage}
\\

\vspace{0.1cm}
\begin{minipage}{0.32\linewidth}
\centerline{\footnotesize{(a)}}
\end{minipage}
\begin{minipage}{0.32\linewidth}
\centerline{\footnotesize{(b)}}
\end{minipage}
\begin{minipage}{0.32\linewidth}
\centerline{\footnotesize{(c)}}
\end{minipage}
\\

\vspace{0.1cm}
\begin{minipage}{0.32\linewidth}
\centerline{\includegraphics[scale=0.30]{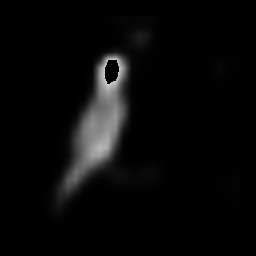}}
\end{minipage}
\begin{minipage}{0.32\linewidth}
\centerline{\includegraphics[scale=0.30]{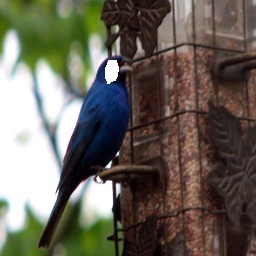}}
\end{minipage}
\begin{minipage}{0.32\linewidth}
\centerline{\includegraphics[scale=0.30]{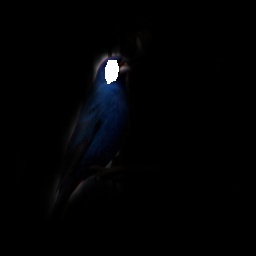}}
\end{minipage}
\\

\vspace{0.1cm}
\begin{minipage}{0.32\linewidth}
\centerline{\footnotesize{(d)}}
\end{minipage}
\begin{minipage}{0.32\linewidth}
\centerline{\footnotesize{(e)}}
\end{minipage}
\begin{minipage}{0.32\linewidth}
\centerline{\footnotesize{(f)}}
\end{minipage}
   \caption{Example of adversarial erasing. (a) Input image; (b) Predicted foreground map; (c) Adversarially erased binary mask; (d) Adversarially erased foreground map; (e) Adversarially erased feature map; (f) Adversarially erased foreground feature map. While we apply adversarial erasing to feature maps from the classifier rather than the image itself, the erased regions are overlaid on the image and denoted by the white color for better visibility.}
\label{fig:adv_erasing}
\end{center}
\end{figure}

\fref{fig:adv_erasing} visualizes an example of adversarial erasing. Given the predicted foreground probability map in (b), we erase the high activation regions to eliminate the dependency on them. While we apply adversarial erasing to feature maps, we visualize its result on the images for better visibility. (c) and (e) illustrate the case when the classifier is forced to learn classification without relying on the most discriminative region. The corresponding loss is naturally propagated through the classifier and the feature extractor, similar to $\calL_{cls}$. (d) and (f) illustrate the case when the localizer and the classifier work together to activate the object region and classify the region correctly, without the discriminative region. Therefore, in this case, the associated loss is used to update not only the classifier and the feature extractor but also the localizer, similar to $\calL_{cls\text{-}fg}$.

$\calL_{ae}$ corresponds to the case of (c) and (e) in \fref{fig:adv_erasing}. It is to reduce the dependency of the classification on a highly discriminative region (high score region on the score map $\mF^c$) and increase classification scores of less discriminative regions of the object. As the network is trained by using only image-level class label, the trained network typically focuses on discriminating differences between classes and discovers only the highly discriminative regions. Accordingly, the learned network often fails to localize an entire object. Hence, the network is trained to depend on not only highly discriminative regions but also other areas to localize better. It is achieved by explicitly disconnecting the dependency on the discriminative parts by removing the corresponding regions on the feature map. 

In detail, a binary mask $\mF^{e_1} \in \mathbb{Z}^{28 \times 28}$ is generated based on the predicted foreground mask $\mF^{fg}$. The binary mask assigns zeros to its elements if their corresponding foreground probabilities exceed a threshold, and positive ones otherwise. Specifically, the mask $\mF^{e_1}$ is generated as follows:
\begin{equation}
	\mF^{e_1}_{i,j} = \left\{
	\begin{array}{l l}
	0, \; \text{ if } \mF^{fg}_{i,j} \geq t_1  \\
	1, \; \text{ otherwise } \\ 
	\end{array} \right.
\label{eq:erased_map_1}
\end{equation}
where $\mF^{fg}_{i,j}$ denotes the predicted foreground probability at ($i,j$), and $t_1$ represents a threshold. This process is denoted by $E_{1}(\cdot)$ in~\fref{fig:framework_training}. Subsequently, the mask $\mF^{e_1}$ is downsampled and multiplied with the score map $\mF^c$, similar to~\eref{eq:loss_foreground}. This binary mask replaces elements with high foreground probabilities with zeros while retaining the original scores for other elements when multiplied element-wise. The threshold $t_1$ determines the extent of adversarial erasing where a lower threshold corresponds to more pronounced erasing.

The output of multiplication is then processed by global average pooling and a softmax function. Finally, the adversarially erased classification loss $\calL_{ae}$ is computed using a cross-entropy loss function.
\begin{equation}
\begin{split}
	\calL_{ae} = - \sum_{i=1}^{C} \vy_i  \ln \big( \sigma ( \text{GAP} ( \mF^c  \cdot  \hat{\mF}^{e_1} ))_{i} \big)
\end{split}
\label{eq:loss_erased_binary}
\end{equation}
where $\hat{\mF}^{e_1}$ denotes the downsampled map of $\mF^{e_1}$. The process of multiplying this mask to $\mF^c$ is denoted by $E_1(\cdot)$ in~\fref{fig:framework_training}.

\noindent \textbf{Adversarially erased foreground classification loss $\calL_{ae\text{-}fg}$.}
Similar to $\calL_{ae}$, it is to alleviate the dependency on a highly discriminative region and to localize an entire object. However, different from $\calL_{ae}$, it is to guide both classifier and localizer like $\calL_{cls\text{-}fg}$. Therefore, $\calL_{ae\text{-}fg}$ corresponds to the case of (d) and (f) in \fref{fig:adv_erasing}. While $\calL_{ae}$ utilizes $\mF^{e_1}$ that propagates all elements equally except highly discriminative parts, this loss $\calL_{cls\text{-}fg}$ uses the predicted foreground map to control pixel-level information flow of the score map $\mF^c$. Hence, it affects the training of the localizer as well as the classifier. 

In detail, given the foreground mask $\mF^{fg}$, an erased foreground mask $\mF^{e_2} \in \mathbb{R}^{28 \times 28}$ is generated by replacing elements with high values by zeros. This erasing function is denoted by $E_2(\cdot)$ in~\fref{fig:framework_training} and the mask $\mF^{e_2}$ is generated as follows:
\begin{equation}
	\mF^{e_2}_{i,j} = \left\{
	\begin{array}{l l}
	0, \;  & \text{ if } \mF^{fg}_{i,j} \geq t_2  \\
	\mF^{fg}_{i,j}, \; & \text{ otherwise } \\ 
	\end{array} \right.
\label{eq:erased_map_2}
\end{equation}
Similar to~\eref{eq:loss_erased_binary}, after following operations, the adversarially erased foreground loss $\calL_{ae\text{-}fg}$ is computed by using a cross-entropy loss function.
\begin{equation}
\begin{split}
	\calL_{ae\text{-}fg} = - \sum_{i=1}^{C} \vy_i  \ln \big( \sigma ( \text{GAP} ( \mF^c  \cdot  \hat{\mF}^{e_2} ))_{i} \big)
\end{split}
\label{eq:loss_erased_foreground}
\end{equation}
where $\hat{\mF}^{e_2}$ denotes the downsampled map of $\mF^{e_2}$.

\begin{figure}[!t] 
\begin{center}
\begin{minipage}{0.32\linewidth}
\centerline{\includegraphics[scale=0.3]{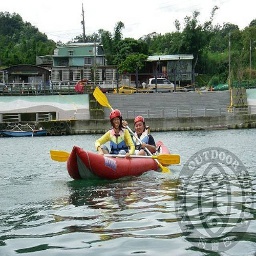}}
\end{minipage}
\begin{minipage}{0.32\linewidth}
\centerline{\includegraphics[scale=0.3]{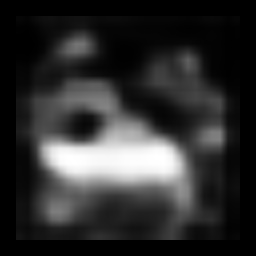}}
\end{minipage}
\begin{minipage}{0.32\linewidth}
\centerline{\includegraphics[scale=0.3]{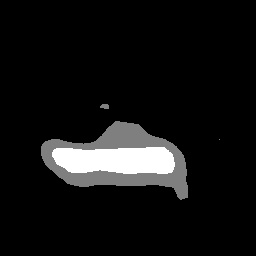}}
\end{minipage}
\\

\vspace{0.1cm}
\begin{minipage}{0.32\linewidth}
\centerline{\footnotesize{(a)}}
\end{minipage}
\begin{minipage}{0.32\linewidth}
\centerline{\footnotesize{(b)}}
\end{minipage}
\begin{minipage}{0.32\linewidth}
\centerline{\footnotesize{(c)}}
\end{minipage}
   \caption{Example of pseudo-label generation. (a) Input image; (b) Predicted foreground probability map; (c) Pixel-level pseudo labels. In (c), white and black denote pseudo foreground and background regions, respectively, while gray represents uncertainty. The gray region is not used for loss computation.}
\label{fig:pseudo_label_generation}
\end{center}
\end{figure}

\noindent \textbf{Pseudo label loss $\calL_{pseudo}$.}
Due to the lack of detailed location information for training the localizer, we generate pseudo labels and utilize them to guide the localizer to suppress activation values in the background and increase them in the foreground. \cite{Zhang2020} and~\cite{Guo2021} generated pseudo bounding box labels using a pre-trained network and its localizer, trained a separate bounding box regressor, and used it during inference. In contrast, we generate pixel-level pseudo labels and employ a single shared feature extractor for both training and inference.

\fref{fig:pseudo_label_generation} shows the pseudo-label generation process. Given the predicted foreground probability map, regions with high and low activation are identified as the pseudo foreground and background, respectively, while the intermediate region is considered uncertain. In (c), the pseudo foreground, background, and uncertain regions are represented by white, black, and gray, respectively. Then, the localizer is trained to suppress activation values in the pseudo background to zero and to increase them in the pseudo foreground to one. The uncertain regions are used for training.

$\calL_{pseudo}$ is to explicitly guide the localizer by using pixel-level pseudo labels. Assuming the pseudo labels are reasonable, it is used to either increase or decrease the activation values. For the elements that the pseudo labels believe an object exists at, the loss aims to increase the corresponding activation values. The loss obviously aims to reduce the activation values of the elements that the pseudo labels believe the object is not at. To reduce incorrect guidance, it ignores elements with an intermediate value since it is challenging to determine correct pseudo labels. 

In detail, a pseudo label map $\mF^{psd}$ is generated using the foreground mask $\mF^{fg}$. The map $\mF^{psd}$ assigns positive ones to its elements if their corresponding foreground probabilities exceed a threshold $t_3$, and assigns zeros if they are below another threshold $t_4$ where $t_4 < t_3$. Specifically, the pseudo label map $\mF^{psd}$ is generated as follows:
\begin{equation}
	\mF^{psd}_{i,j} = \left\{
	\begin{array}{l l}
	1, \; \text{ if } \mF^{fg}_{i,j} \geq t_3  \\
	0, \; \text{ if } \mF^{fg}_{i,j} \leq t_4  \\
	\end{array} \right.
\label{eq:pseudo_GT}
\end{equation}
where $\mF^{fg}_{i,j}$ denotes the predicted foreground probability at ($i,j$); $t_3$ and $t_4$ represent thresholds for the foreground and background, respectively. Since the elements with intermediate values are ambiguous between foreground and background, pseudo labels are not generated, and the corresponding regions are not used to compute the loss $\calL_{pseudo}$. This process is denoted by $E_{psd}(\cdot)$ in~\fref{fig:framework_training}.

Then, the pseudo label loss $\calL_{pseudo}$ is computed by measuring $\ell_1$ distance between the pseudo label map $\mF^{psd}$ and the predicted foreground mask $\mF^{fg}$. Specifically, we compute the distances for object existing and non-existing regions separately and add them. 
\begin{equation}
\begin{split}
	&\calL_{psd\text{-}fg} = \frac{1}{N} \sum_i \sum_j \big( \mathbbm{1}(\mF^{fg}_{i,j} \geq t_3 ) | \mF^{psd}_{i,j} - \mF^{fg}_{i,j} |  \big) \\
	&\calL_{psd\text{-}bg} = \frac{1}{N} \sum_i \sum_j \big( \mathbbm{1}(\mF^{fg}_{i,j} \leq t_4) | \mF^{psd}_{i,j} - \mF^{fg}_{i,j} |  \big) \\
	&\calL_{pseudo} = \calL_{psd\text{-}fg} + \calL_{psd\text{-}bg}
\end{split}
\label{eq:loss_pseudo}
\end{equation}
where $\mathbbm{1}(\cdot)$ denotes an indicator function that returns one when the argument is true and zero otherwise; $N$ denotes the total number of elements of the map.

\noindent \textbf{Background activation suppression loss $\calL_{bas}$.}
It is to guide the network to predict higher classification score for foreground and lower value for background. To achieve this, it compares the average score of the entire image and that of the background region. 

Specifically, given the score map $\mF^c$ from the classifier, the score map for the ground-truth class is obtained by extracting the channel that corresponds to the ground-truth class. It is then processed by a global average pooling to obtain the average score $s_{all} \in \mathbb{R}$ from the entire image. It is computed as follows: 
\begin{equation}
\begin{split}
	s_{all} = \text{GAP}(\mF^c_{\vy_n})
\end{split}
\label{eq:s_all}
\end{equation}
where $\vy_n$ denotes the ground-truth class. Note that the order of GAP and slice is interchangeable. 

To obtain the average score $s_{bg}$ from the background region, the feature map $\mF^f$ from the feature extractor is first multiplied with the negated foreground mask $\mF^{fg}$. The multiplication is processed by an element-wise multiplication for each channel. This process is denoted by $E_3(\cdot)$ in~\fref{fig:framework_training}. The output of the multiplication is then processed by the classifier to obtain the score map from the background region. Then, similar to~\eref{eq:s_all}, the map that corresponds to the ground-truth class is extracted and processed by a global average pooling. 
\begin{equation}
\begin{split}	
	s_{bg} = \text{GAP}\big(E_c(\mF^f \cdot (1-\mF^{fg}))_{\vy_n}\big)
\end{split}
\label{eq:s_bg}
\end{equation}

Finally, the background activation suppression loss $\calL_{bas}$ is computed as follows:
\begin{equation}
\begin{split}
	\calL_{bas} = \frac{s_{bg}}{s_{all} + \epsilon} 
\end{split}
\label{eq:loss_BAS}
\end{equation}
Exceptionally, when $s_{bg}$ is larger than $s_{all}$, the loss is ignored to avoid unstable training such as incorrectly predicted foreground mask.

\noindent \textbf{Area constraint loss $\calL_{ac}$.}
It is to limit the region of the foreground. Considering $\calL_{bas}$, increasing the values in $\mF^{fg}$ obviously decreases $s_{bg}$ and $\calL_{bas}$. Hence, to avoid having large values for all the elements in $\mF^{fg}$, the area constraint loss $\calL_{ac}$ is utilized.

Given the foreground mask $\mF^{fg}$, an average over the spatial dimension is computed and used as the area constraint loss $\calL_{ac}$.
\begin{equation}
\begin{split}
	\calL_{ac} = \frac{1}{N} \sum_i \sum_j \mF^{fg}_{i,j}
\end{split}
\label{eq:loss_AC}
\end{equation}
where $N$ denotes the total number of elements in $\mF^{fg}$.

\noindent \textbf{Total loss.}
Finally, it is computed by a weighted summation of $\calL_{cls}$, $\calL_{cls\text{-}fg}$, $\calL_{ae}$, $\calL_{ae\text{-}fg}$, $\calL_{pseudo}$ $\calL_{bas}$, and $\calL_{ac}$. 
\begin{equation}
\begin{split}
	\calL = &\calL_{cls} + \gamma_1 \calL_{cls\text{-}fg} + \gamma_2 \calL_{ae} + \gamma_3 \calL_{ae\text{-}fg} \\
	& + \gamma_4 \calL_{pseudo} + \gamma_5 \calL_{bas} + \gamma_6 \calL_{ac} 
\end{split}
\label{eq:total_loss}
\end{equation}
where $(\gamma_1, \gamma_2, \gamma_3, \gamma_4, \gamma_5, \gamma_6)$ are weighting coefficients to balance the seven loss terms. The weighting coefficients in the total loss are experimentally determined similar to others~\citep{Wu_2022_CVPR, Pan2021}.

\section{EXPERIMENTS AND RESULTS}
\label{sec:results}

\subsection{Experimental Setting}
\label{sec:exp_setting}
\noindent \textbf{Dataset.}
To demonstrate the effectiveness of the proposed method, we experiment using three publicly available datasets (ILSVRC-2012~\citep{ILSVRC15}, CUB-200-2011~\citep{cub_dataset}, and PASCAL VOC 2012~\citep{pascalvoc}). As the ILSVRC-2012 and CUB-200-2011 datasets have been utilized in previous weakly supervised object localization literatures~\citep{Wu_2022_CVPR}, we follow experimental settings of them. In detail, the CUB-200-2011 dataset contains 11,788 images of 200 species of birds~\citep{cub_dataset}. They are split into 5,994 images for training and 5,794 images for testing. The ILSVRC-2012 dataset contains images of 1,000 classes~\citep{ILSVRC15} where the images are split into 1,281,167 for training and 50,000 for testing. We additionally experiment using the PASCAL VOC 2012 classification dataset that contains 5,717 images for training and 5,823 images for validation.



\noindent \textbf{Evaluation metrics.}
First of all, three metrics are utilized to evaluate the performance of each method on a dataset following previous literatures~\citep{Meng2021, Wu_2022_CVPR}. Among three metrics, two of them (Top-1 and Top-5) evaluate classification and localization results together. The other (GT-known) evaluates only localization accuracy by assuming that the prediction for classification is known or correct. Each metric measures the portion of the images that are correctly predicted by computing the Intersection over Union (IoU). In detail, GT-known considers that the prediction is correct if the computed IoU is greater than 0.5. Also, for Top-1 and Top-5, the computed IoU should be greater than 0.5. Additionally, for Top-1, the prediction for classification should match the ground-truth class. For Top-5, if the ground-truth class is one of the five classes with the highest scores for classification, this metric considers that the prediction for classification is correct.

Additionally, we propose to utilize mean Intersection over Union (mIoU) to evaluate localization accuracy in detail. The usual three metrics use a binary measurement that considers the localization is correct if the corresponding IoU is greater than 0.5. Accordingly, the previous metrics do not differentiate the IoU of 0.6 and that of 0.8. Hence, we employ mIoU that does not binarize the IoU measurement and can better evaluate localization performance in detail. The mIoU is obtained by first computing the average IoU of images in each class and then calculating the mean of the average IoU over all the classes.

\noindent \textbf{Optimization.}
The proposed method is trained by using two different backbones, MobileNetV1~\citep{mobilenetv1} and InceptionV3~\citep{inceptionv3}. Both networks are trained by using a stochastic gradient descent optimizer with an initial learning rate of 0.001 and a momentum of 0.9. The networks for the ILSVRC-2012 dataset and the CUB-200-2011 dataset are trained for 9 epochs and 100 epochs, respectively. This is because the ILSVRC-2012 dataset~\citep{ILSVRC15} is much larger than the CUB-200-2011 dataset~\citep{cub_dataset}. 

\noindent \textbf{Implementation.}
The algorithms are implemented by using PyTorch 1.8, OpenCV 4.6, and CUDA 11.1 in Ubuntu 18.04. The networks are trained by using either two NVIDIA RTX A5000 GPU or two NVIDIA GeForce RTX 3090 GPU for the ILSVRC-2012 dataset~\citep{ILSVRC15}. They are trained by using only one GPU for the CUB-200-2011 dataset~\citep{cub_dataset} since this dataset is smaller. The code will be available upon publication.

\subsection{Results}
We present quantitative comparisons on the CUB-200-2011 dataset~\citep{cub_dataset} using the MobileNetV1~\citep{mobilenetv1} and InceptionV3~\citep{inceptionv3} backbones in Tables~\ref{tab:result_cub_mobilenet} and~\ref{tab:result_cub_inception}, respectively. In~\tref{tab:result_cub_mobilenet}, the proposed method is compared with six different methods and achieves state-of-the-art performance in all metrics. \tref{tab:result_cub_mobilenet} demonstrates that the proposed method outperforms 11 previous works in all metrics. Specifically, for the MobileNetV1 backbone, the proposed method outperforms the previous state-of-the-art method by absolute 1.51\%, 1.00\%, and 0.68\% in Top-1, Top-5, and GT-known localization accuracy, respectively. For the InceptionV3 backbone, the proposed method outperforms by absolute 0.50\%, 1.68\%, and 1.84\%, respectively. The results of the previous methods are from the original papers.

\begin{table}[!t]
\small
\centering
\begin{minipage}{0.99\linewidth}
\caption{Quantitative comparison of Top-1, Top-5, and GT-known localization accuracies using MobileNetV1 backbone~\citep{mobilenetv1} on the CUB-200-2011 dataset~\citep{cub_dataset}.}
\label{tab:result_cub_mobilenet}
\renewcommand{\arraystretch}{1.} 
\centering
\begin{tabular}{ >{\centering}m{0.47\textwidth}| *{2}{>{\centering}m{0.11\textwidth}|} >{\centering\arraybackslash}m{0.11\textwidth} } 
\hline
Method & Top-1 & Top-5 & GT-known  \\ 
\hline\hline
CAM~\citep{cam2016} & 48.07 & 59.20 & 63.30 \\
HaS~\citep{hideseek2017} & 46.70 & - & 67.31   \\
ADL~\citep{Choe2019} & 47.74 & - & -  \\
RCAM~\citep{Bae2020} & 59.41 & - & 78.60   \\
FAM~\citep{Meng2021} & 65.67 & - & 85.71   \\
BAS~\citep{Wu_2022_CVPR} & \underline{69.77} & \underline{86.00} & \underline{92.35} \\
Proposed & \textbf{71.28} & \textbf{87.00} & \textbf{93.03} \\
\hline  
\end{tabular}
\end{minipage}
\end{table}

\begin{table}[!t]
\small
\centering
\begin{minipage}{0.99\linewidth}
\caption{Quantitative comparison of Top-1, Top-5, and GT-known localization accuracies using InceptionV3 backbone~\citep{inceptionv3} on the CUB-200-2011 dataset~\citep{cub_dataset}.}
\label{tab:result_cub_inception}
\renewcommand{\arraystretch}{1.} 
\centering
\begin{tabular}{ >{\centering}m{0.47\textwidth}| *{2}{>{\centering}m{0.11\textwidth}|} >{\centering\arraybackslash}m{0.11\textwidth} } 
\hline
Method & Top-1 & Top-5 & GT-known  \\ 
\hline\hline
CAM~\citep{cam2016} & 41.06 & 50.66 & 55.10   \\
SPG~\citep{spg2018} & 46.64 & 57.72 & -   \\
DANet~\citep{Xue2019} & 49.45 & 60.46 & 67.03  \\
PSOL~\citep{Zhang2020} &  65.51 & 83.44 & -   \\
I2C~\citep{i2c2020} & 55.99 & 68.34 & 72.60   \\
GCNet~\citep{Lu2020} & 58.58 & 71.00 & 75.30  \\
RCAM~\citep{Bae2020} & 53.04 & - & 69.95  \\
SPA~\citep{Pan2021} & 53.59 & 66.50 & 72.14   \\
SLT~\citep{Guo2021} &  66.10 & - & 86.50  \\
FAM~\citep{Meng2021} & 70.67 & - & 87.25  \\
BAS~\citep{Wu_2022_CVPR} &\underline{73.29}&\underline{86.31}&\underline{92.24} \\
Proposed & \textbf{73.79}& \textbf{87.99}&\textbf{94.08} \\
\hline  
\end{tabular}
\end{minipage}
\end{table}

Tables~\ref{tab:result_ilsvrc_mobilenet} and~\ref{tab:result_ilsvrc_inception} present the quantitative results on the ILSVRC-2012 dataset~\citep{ILSVRC15} using the MobileNetV1~\citep{mobilenetv1} and InceptionV3~\citep{inceptionv3} backbones, respectively. In~\tref{tab:result_ilsvrc_mobilenet}, the proposed method is compared with six previous works and achieves the best performance in all metrics. \tref{tab:result_ilsvrc_inception} verifies that the proposed method outperforms the previous 11 methods in all metrics. Specifically, for the MobileNetV1 backbone, the proposed method achieves absolute 0.77\%, 0.52\%, and 0.31\% higher accuracy for Top-1, Top-5, and GT-known metrics, respectively. For the InceptionV3 backbone, the proposed method achieves absolute 0.34\%, 0.50\%, and 0.59\% higher accuracy, respectively.

\begin{table}[!t]
\small
\centering
\begin{minipage}{0.99\linewidth}
\caption{Quantitative comparison of Top-1, Top-5, and GT-known localization accuracies using MobileNetV1 backbone~\citep{mobilenetv1} on the ILSVRC-2012 dataset~\citep{ILSVRC15}.}
\label{tab:result_ilsvrc_mobilenet}
\renewcommand{\arraystretch}{1.} 
\centering
\begin{tabular}{ >{\centering}m{0.47\textwidth}| *{2}{>{\centering}m{0.11\textwidth}|} >{\centering\arraybackslash}m{0.11\textwidth} } 
\hline
Method & Top-1 & Top-5 & GT-known \\ 
\hline\hline
CAM~\citep{cam2016} &  43.35 & 54.44 & 58.97 \\
HaS~\citep{hideseek2017} & 42.73 & - & 60.12 \\
ADL~\citep{Choe2019} & 43.01 & - & - \\
RCAM~\citep{Bae2020} & 44.78 & - & 61.69 \\
FAM~\citep{Meng2021} & 46.24 & - & 62.05 \\
BAS~\citep{Wu_2022_CVPR} & \underline{52.97}& \underline{66.59}&\underline{72.00}\\
Proposed & \textbf{53.74}& \textbf{67.11} & \textbf{72.31}\\
\hline  		
\end{tabular}
\end{minipage}
\end{table}

\begin{table}[!t]
\small
\centering
\begin{minipage}{0.99\linewidth}
\caption{Quantitative comparison of Top-1, Top-5, and GT-known localization accuracies using InceptionV3 backbone~\citep{inceptionv3} on the ILSVRC-2012 dataset~\citep{ILSVRC15}.}
\label{tab:result_ilsvrc_inception}
\renewcommand{\arraystretch}{1.} 
\centering
\begin{tabular}{ >{\centering}m{0.47\textwidth}| *{2}{>{\centering}m{0.11\textwidth}|} >{\centering\arraybackslash}m{0.11\textwidth} } 
\hline
Method &  Top-1 & Top-5 & GT-known \\ 
\hline\hline
CAM~\citep{cam2016}  &  46.29 & 58.19 & 62.68 \\
SPG~\citep{spg2018}  & 48.60 & 60.00 & 64.69 \\
DANet~\citep{Xue2019}  & 47.53 & 58.28 & - \\
PSOL~\citep{Zhang2020} &  54.82 & 63.25 & 65.21 \\
I2C~\citep{i2c2020} & 53.11 & 64.13 & 68.50 \\
GCNet~\citep{Lu2020} & 49.06 & 58.09 & - \\
RCAM~\citep{Bae2020} & 50.56 & - & 64.44 \\
SPA~\citep{Pan2021} & 52.73 & 64.27 & 68.33 \\
SLT~\citep{Guo2021} &  55.70 & 65.40 & 67.60 \\
FAM~\citep{Meng2021} & 55.24 & - &68.62\\
BAS~\citep{Wu_2022_CVPR} &\underline{58.51}&\underline{69.00}&\underline{71.93}\\
Proposed & \textbf{58.85}& \textbf{69.50}&\textbf{72.52}\\
\hline  
\end{tabular}
\end{minipage}
\end{table}

\begin{table}[!t]
\small
\centering
\begin{minipage}{0.97\linewidth}
\caption{Quantitative results on localization accuracy by using mean Intersection over Union (IoU).}
\label{tab:result_iou}
\renewcommand{\arraystretch}{1.} 
\centering
\begin{tabular}{ >{\centering}m{0.37\textwidth}| >{\centering}m{0.17\textwidth}| >{\centering}m{0.13\textwidth}| >{\centering\arraybackslash}m{0.13\textwidth} } 
\hline
Method & Backbone & CUB & ILSVRC \\ 
\cline{3-4}
\hline\hline
BAS~\citep{Wu_2022_CVPR} & MobileNet & 54.51 & 46.63 \\
Proposed & V1 & \textbf{56.08}& \textbf{46.66} \\
\hline  		

BAS~\citep{Wu_2022_CVPR} & Inception & 52.71 & 48.31 \\
Proposed & V3 & \textbf{53.94}& \textbf{49.98} \\
\hline  
\end{tabular}
\end{minipage}
\end{table}

Additionally, we evaluate localization accuracy using mean Intersection over Union (mIoU) in~\tref{tab:result_iou}. Since we propose to use this metric in addition to the usual three metrics in~\tref{tab:result_cub_inception}, results of previous methods are not reported in previous literature. Since BAS~\citep{Wu_2022_CVPR} achieves the second-best accuracy in all metrics for all datasets and backbones (see Tables~\ref{tab:result_cub_mobilenet},~\ref{tab:result_cub_inception},~\ref{tab:result_ilsvrc_mobilenet}, and~\ref{tab:result_ilsvrc_inception}), we only compare the proposed method with BAS for this additional metric. The results of the BAS are obtained by only processing evaluation where the trained model is provided by the authors of BAS~\citep{Wu_2022_CVPR}. The proposed method achieves an absolute 1.57\% and 1.23\% higher mIoU for the CUB-200-2011 dataset using the MobileNetV1 and InceptionV3 backbones, respectively. It also achieves an absolute 0.03\% and 1.67\% higher mIoU for the ILSVRC-2012 dataset. These results further demonstrate that the proposed method outperforms the previous state-of-the-art method in terms of localization accuracy.

Moreover, we present a quantitative comparison using the PASCAL VOC 2012 classification dataset~\citep{pascalvoc} in Tables~\ref{tab:result_voc_mobilenet} and~\ref{tab:result_voc_inception}. Since this dataset has not been utilized for weakly supervised object localization, results of previous methods are not available in previous literature. Since BAS~\citep{Wu_2022_CVPR} achieves the second-best accuracy in all metrics in Tables~\ref{tab:result_cub_mobilenet},~\ref{tab:result_cub_inception},~\ref{tab:result_ilsvrc_mobilenet}, and~\ref{tab:result_ilsvrc_inception}, we compare the proposed method with BAS for this dataset. For the MobileNetV1 backbone, the proposed method achieves an absolute 2.27\%, 3.55\%, 3.55\%, and 1.00\% higher accuracy for Top-1, Top-5, GT-known, and mIoU metrics, respectively. For the InceptionV3 backbone, the proposed method achieves an absolute 3.18\%, 1.40\%, 1.65\%, and 3.22\% higher accuracy, respectively.

\begin{table}[!t]
\small
\centering
\begin{minipage}{0.99\linewidth}
\caption{Quantitative comparison of localization accuracies using the MobileNetV1 backbone~\citep{mobilenetv1} on the PASCAL VOC 2012 classification dataset~\citep{pascalvoc}.}
\label{tab:result_voc_mobilenet}
\centering
\begin{tabular}{>{\centering}m{0.36\textwidth}| *{3}{>{\centering}m{0.1\textwidth}|} >{\centering\arraybackslash}m{0.09\textwidth}}
\hline
Method & Top-1 & Top-5 & GT-known & mIoU  \\ 
\hline\hline
BAS~\citep{Wu_2022_CVPR} & 36.32 & 47.34 & 48.56 & 33.63 \\
Proposed & \textbf{38.59} & \textbf{50.89} & \textbf{52.11} & \textbf{34.63} \\
\hline  
\end{tabular}
\end{minipage}
\end{table}

\begin{table}[!t]
\small
\centering
\begin{minipage}{0.99\linewidth}
\caption{Quantitative comparison of localization accuracies using the InceptionV3 backbone~\citep{inceptionv3} on the PASCAL VOC 2012 classification dataset~\citep{pascalvoc}.}
\label{tab:result_voc_inception}
\centering
\begin{tabular}{>{\centering}m{0.36\textwidth}| *{3}{>{\centering}m{0.1\textwidth}|} >{\centering\arraybackslash}m{0.09\textwidth}}
\hline
Method & Top-1 & Top-5 & GT-known & mIoU  \\ 
\hline\hline
BAS~\citep{Wu_2022_CVPR} & 36.32 & 44.93 & 45.55  &  32.48  \\
Proposed & \textbf{39.50} & \textbf{46.33} & \textbf{47.20} & \textbf{35.70} \\
\hline  
\end{tabular}
\end{minipage}
\end{table}

\begin{figure*}[!t] 
\centering
\begin{minipage}{0.32\linewidth}
\centerline{\includegraphics[scale=0.35]{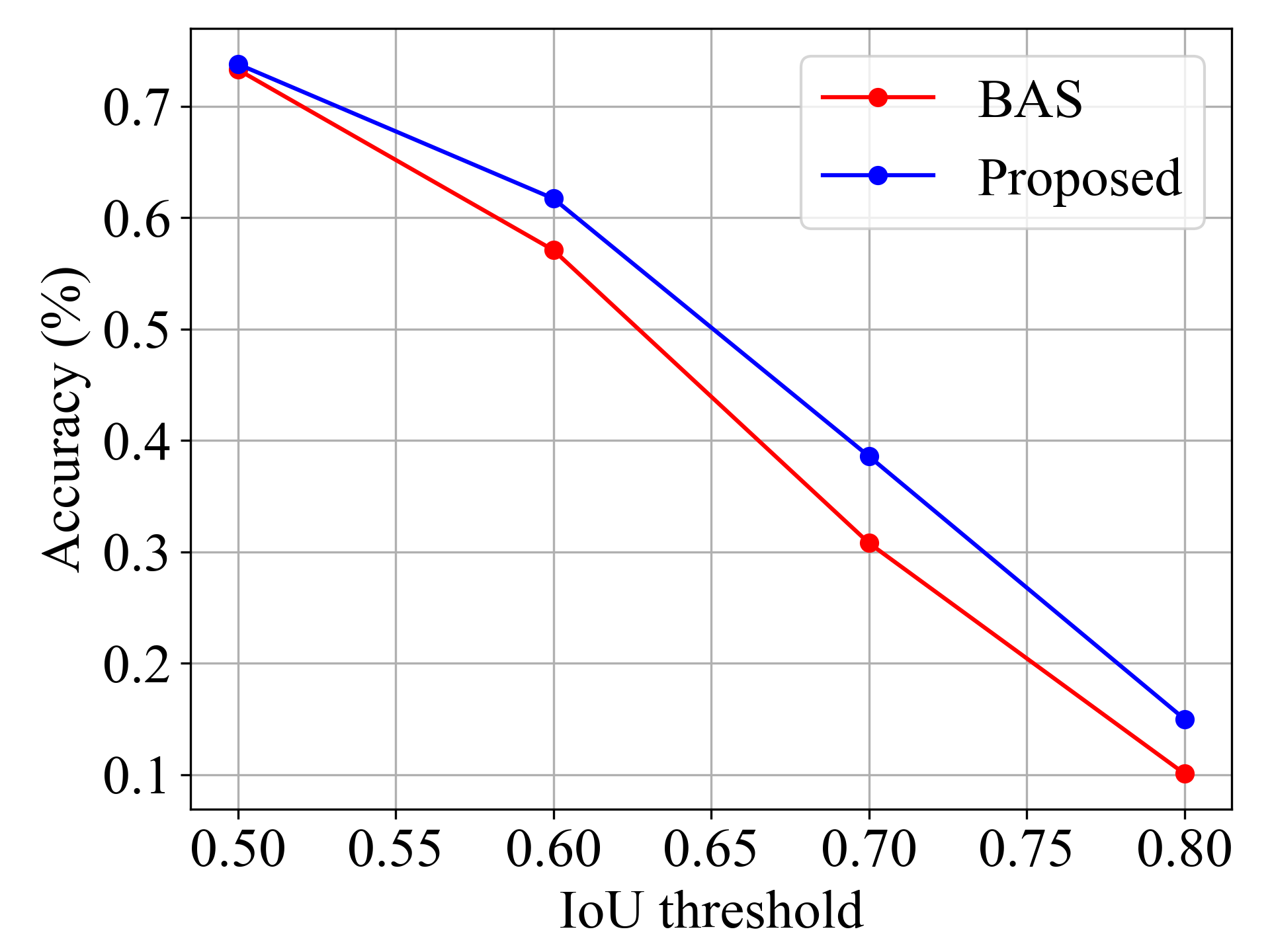}}
\end{minipage}
\begin{minipage}{0.32\linewidth}
\centerline{\includegraphics[scale=0.35]{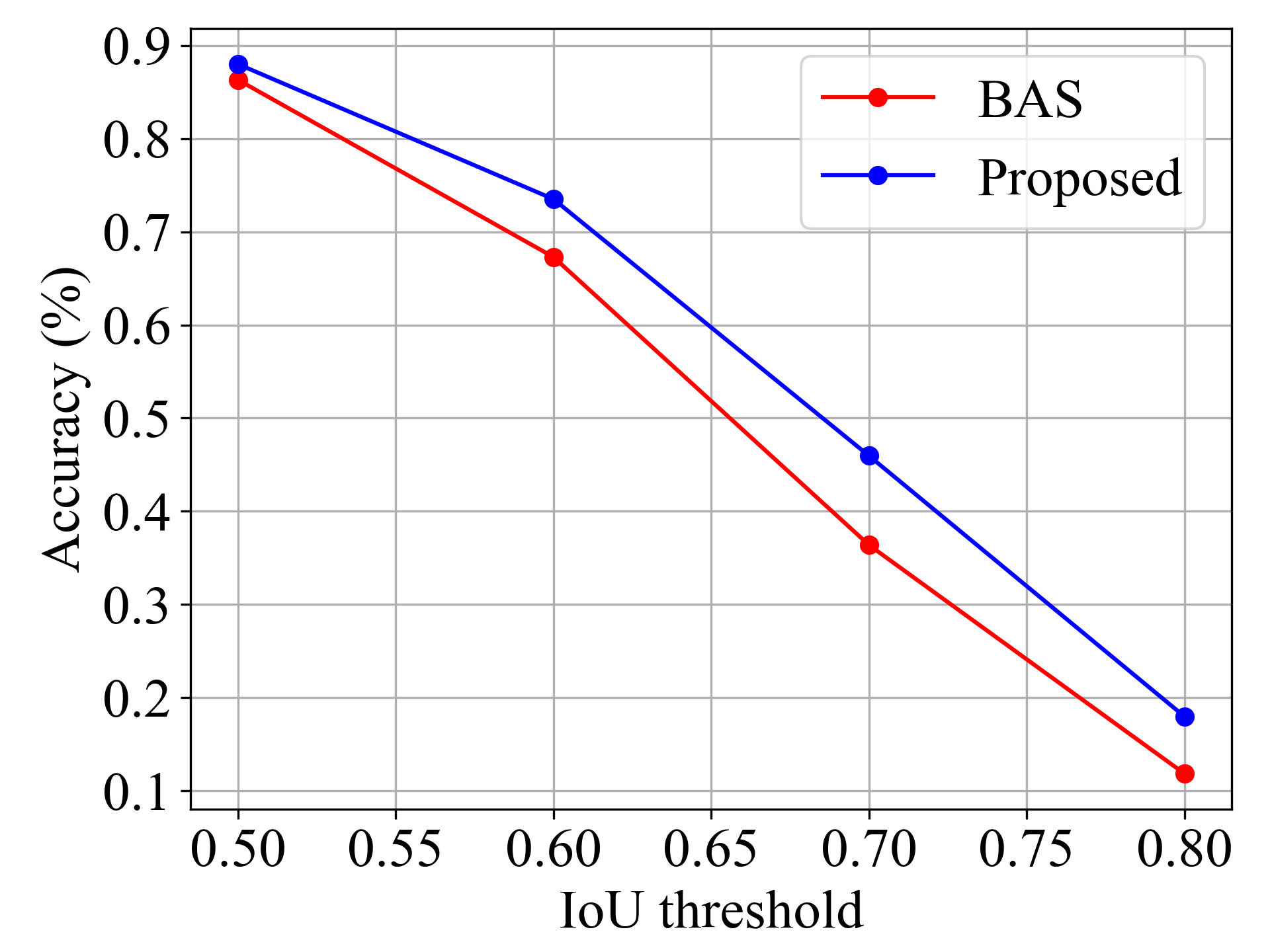}}
\end{minipage}
\begin{minipage}{0.32\linewidth}
\centerline{\includegraphics[scale=0.35]{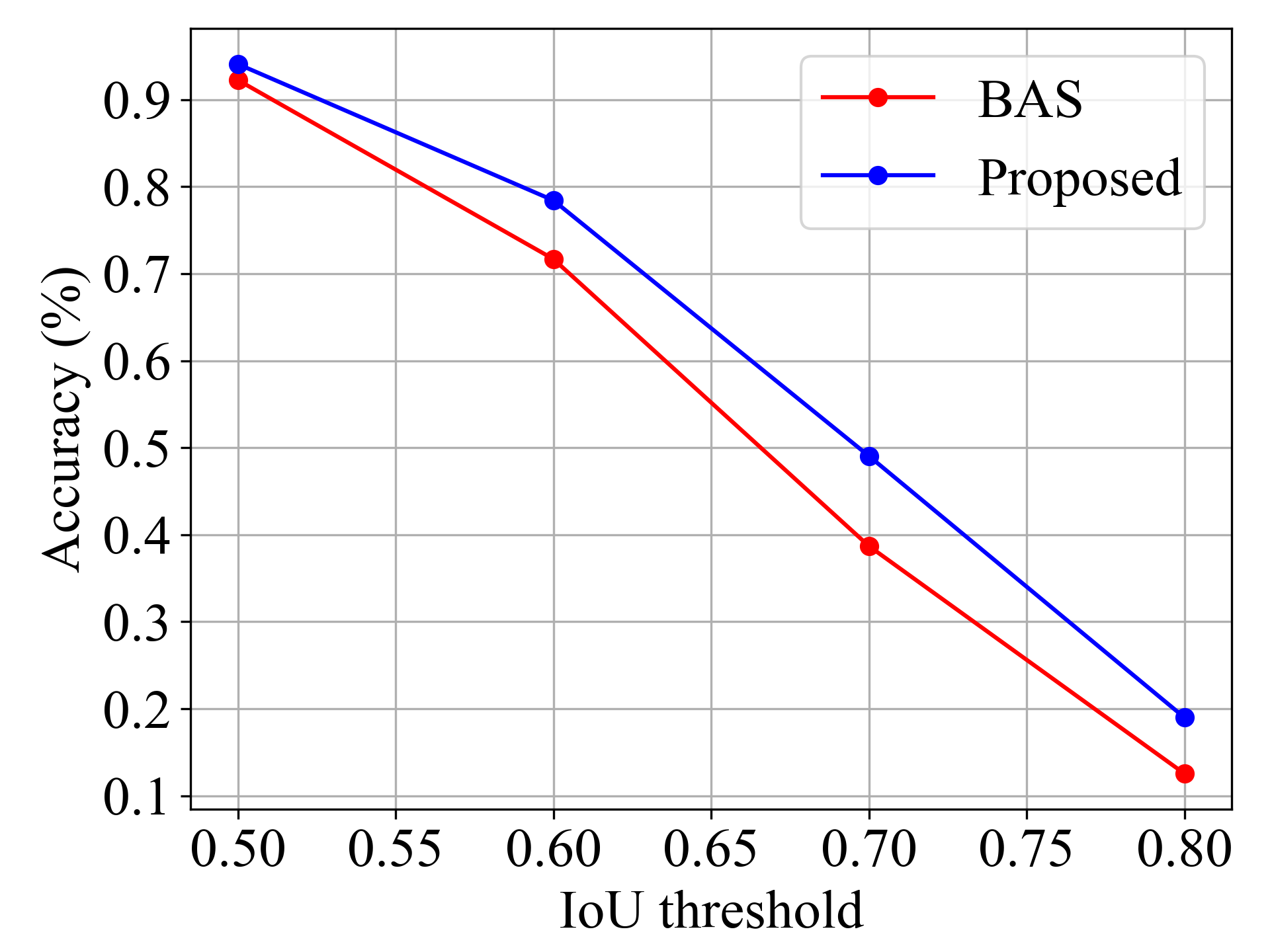}}
\end{minipage}
\\

\vspace{0.1cm}
\begin{minipage}{0.32\linewidth}
\centerline{\footnotesize{(a) Top-1 localization accuracy}}
\end{minipage}
\begin{minipage}{0.32\linewidth}
\centerline{\footnotesize{(b) Top-5 localization accuracy}}
\end{minipage}
\begin{minipage}{0.32\linewidth}
\centerline{\footnotesize{(c) GT-known localization accuracy}}
\end{minipage}
   \caption{Comparison of localization accuracies over varying IoU thresholds between the proposed method and BAS~\citep{Wu_2022_CVPR}. Experiments are conducted using the InceptionV3 backbone~\citep{inceptionv3} on the CUB-200-2011 dataset~\citep{cub_dataset}.}
\label{fig:graph_cub_inception}
\end{figure*}

\begin{figure*}[!t] 
\centering
\begin{minipage}{0.32\linewidth}
\centerline{\includegraphics[scale=0.35]{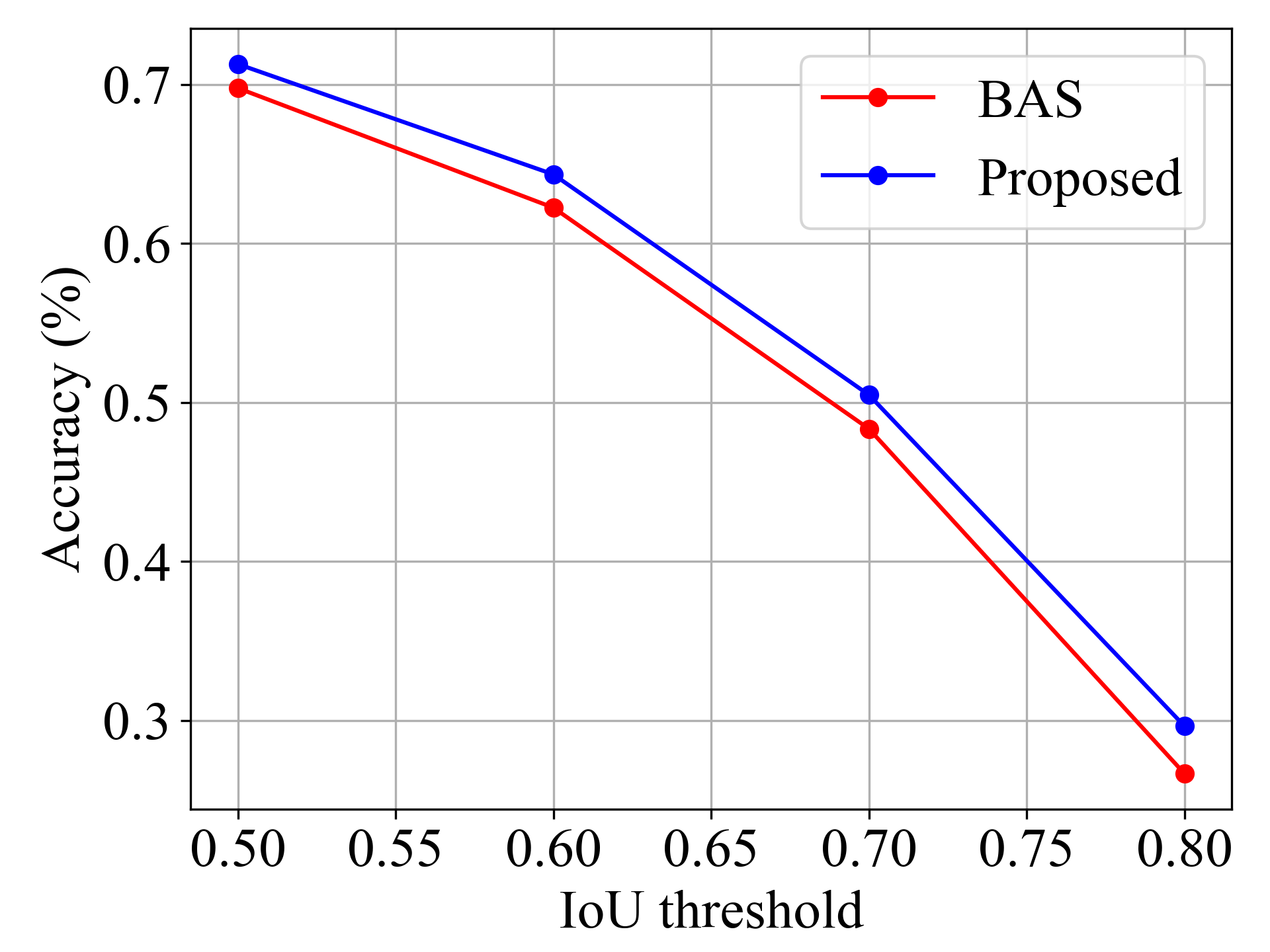}}
\end{minipage}
\begin{minipage}{0.32\linewidth}
\centerline{\includegraphics[scale=0.35]{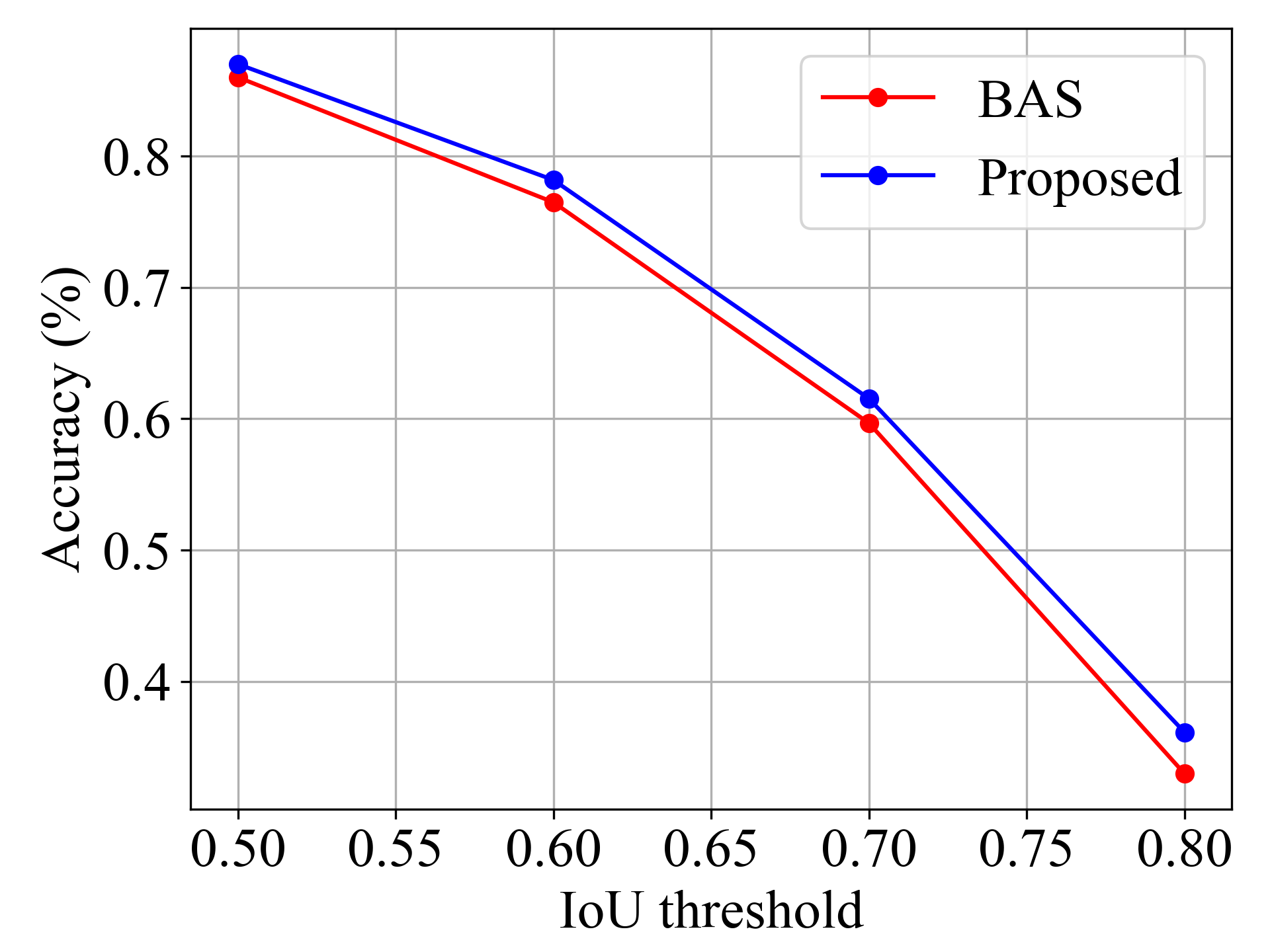}}
\end{minipage}
\begin{minipage}{0.32\linewidth}
\centerline{\includegraphics[scale=0.35]{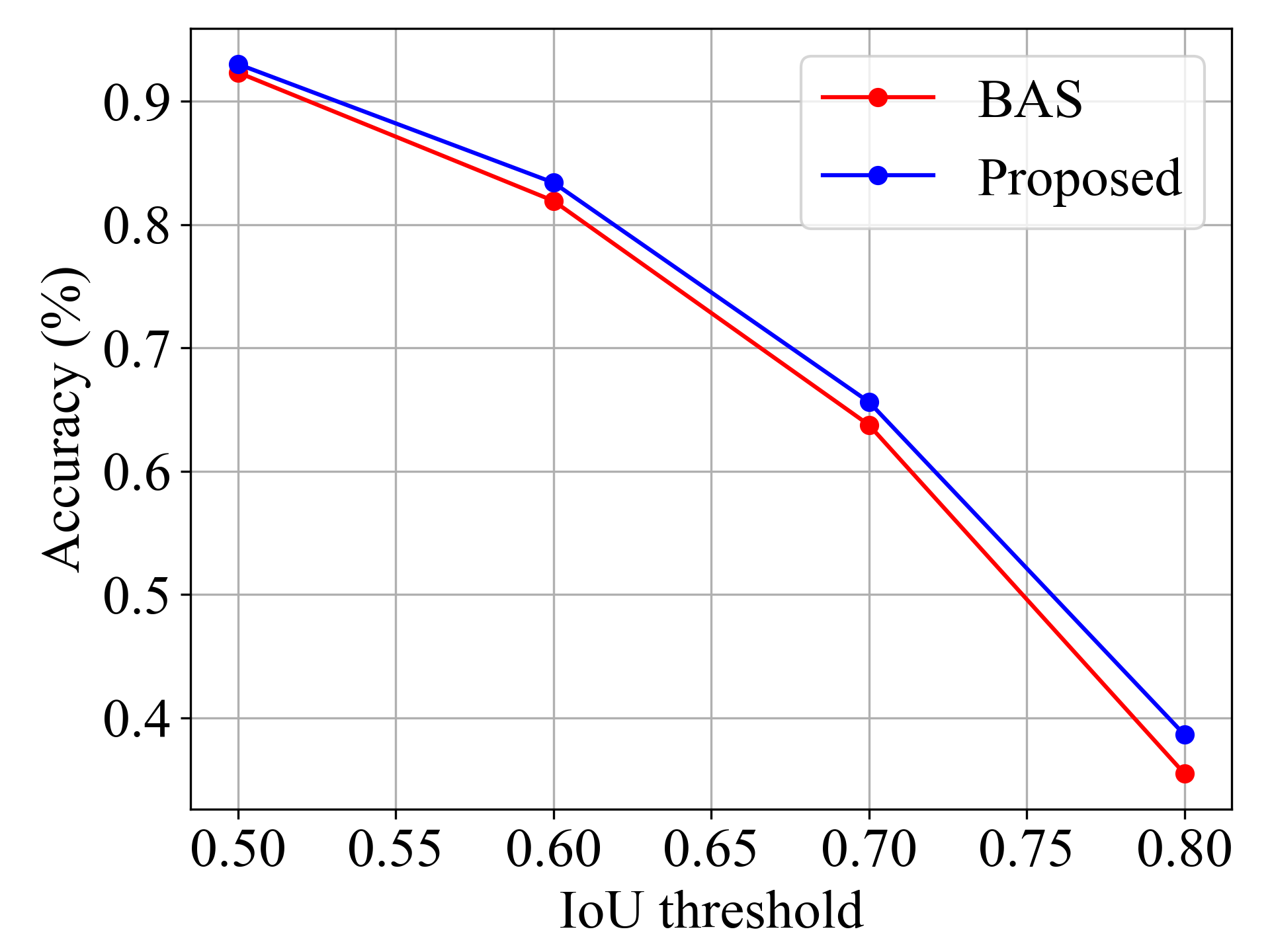}}
\end{minipage}
\\

\vspace{0.1cm}
\begin{minipage}{0.32\linewidth}
\centerline{\footnotesize{(a) Top-1 localization accuracy}}
\end{minipage}
\begin{minipage}{0.32\linewidth}
\centerline{\footnotesize{(b) Top-5 localization accuracy}}
\end{minipage}
\begin{minipage}{0.32\linewidth}
\centerline{\footnotesize{(c) GT-known localization accuracy}}
\end{minipage}
   \caption{Comparison of localization accuracies over varying IoU thresholds between the proposed method and BAS~\citep{Wu_2022_CVPR}. Experiments are performed using the MobileNetV1 backbone~\citep{mobilenetv1} on the CUB-200-2011 dataset~\citep{cub_dataset}.}
\label{fig:graph_cub_mobilenet}
\end{figure*}

Figures~\ref{fig:graph_cub_inception} and~\ref{fig:graph_cub_mobilenet} show the comparison of localization accuracies over varying IoU thresholds between the proposed method and BAS~\citep{Wu_2022_CVPR} which achieves the second-best performance in all metrics. As mentioned in~\sref{sec:exp_setting}, traditional Top-1, Top-5, and GT-known localization accuracy metrics consider that localization is correct if the IoU is greater than 0.5. Accordingly, these metrics have limitations in evaluating localization accuracies in detail. Therefore, we present the localization accuracies over varying IoU thresholds. The experimental results demonstrate that the proposed method consistently outperforms the previous state-of-the-art method~\citep{Wu_2022_CVPR}. The experiments are performed using the InceptionV3 backbone~\citep{inceptionv3} for~\fref{fig:graph_cub_inception} and the MobileNetV1 backbone~\citep{mobilenetv1} for~\fref{fig:graph_cub_mobilenet}. In each figure, (a), (b), and (c) visualize the Top-1, Top-5, and GT-known localization accuracies.

\begin{figure*}[!t] 
\begin{center}
\begin{minipage}{0.03\linewidth}
\centerline{(a)}
\end{minipage}
\begin{minipage}{0.11\linewidth}
\centerline{\includegraphics[scale=0.20]{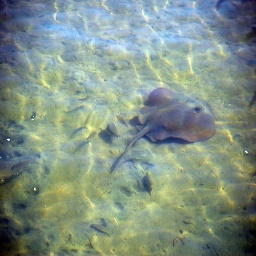}}
\end{minipage}
\begin{minipage}{0.11\linewidth}
\centerline{\includegraphics[scale=0.20]{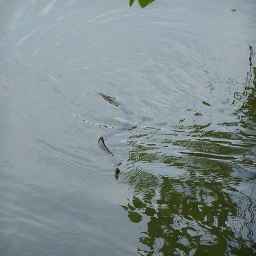}}
\end{minipage}
\begin{minipage}{0.11\linewidth}
\centerline{\includegraphics[scale=0.20]{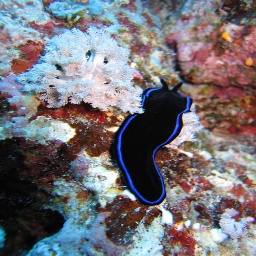}}
\end{minipage}
\begin{minipage}{0.11\linewidth}
\centerline{\includegraphics[scale=0.20]{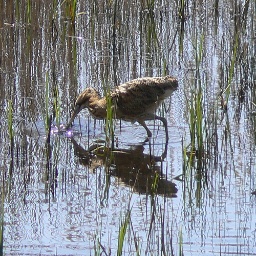}}
\end{minipage}
\begin{minipage}{0.11\linewidth}
\centerline{\includegraphics[scale=0.20]{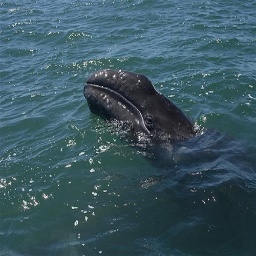}}
\end{minipage}
\begin{minipage}{0.11\linewidth}
\centerline{\includegraphics[scale=0.20]{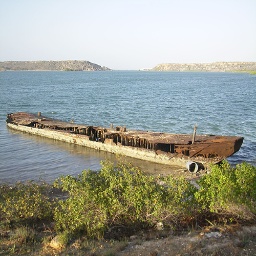}}
\end{minipage}
\\

\vspace{0.1cm}
\begin{minipage}{0.03\linewidth}
\centerline{(b)}
\end{minipage}
\begin{minipage}{0.11\linewidth}
\centerline{\includegraphics[scale=0.20]{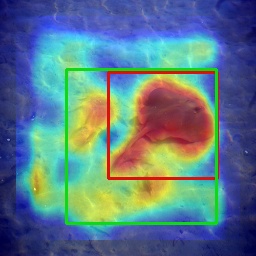}}
\end{minipage}
\begin{minipage}{0.11\linewidth}
\centerline{\includegraphics[scale=0.20]{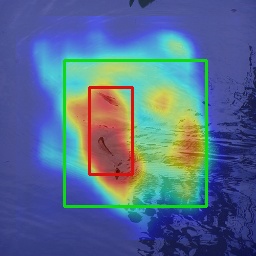}}
\end{minipage}
\begin{minipage}{0.11\linewidth}
\centerline{\includegraphics[scale=0.20]{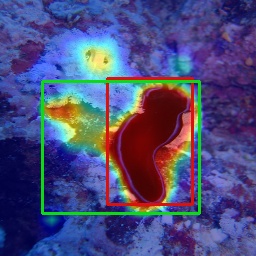}}
\end{minipage}
\begin{minipage}{0.11\linewidth}
\centerline{\includegraphics[scale=0.20]{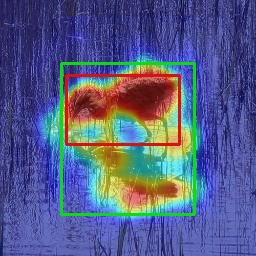}}
\end{minipage}
\begin{minipage}{0.11\linewidth}
\centerline{\includegraphics[scale=0.20]{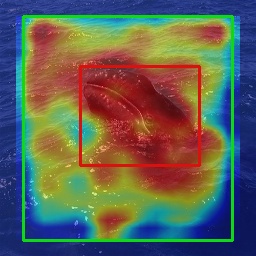}}
\end{minipage}
\begin{minipage}{0.11\linewidth}
\centerline{\includegraphics[scale=0.20]{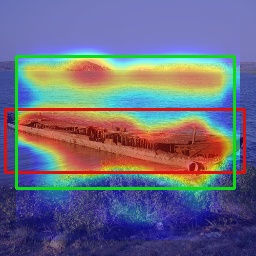}}
\end{minipage}
\\

\vspace{0.1cm}
\begin{minipage}{0.03\linewidth}
\centerline{(c)}
\end{minipage}
\begin{minipage}{0.11\linewidth}
\centerline{\includegraphics[scale=0.20]{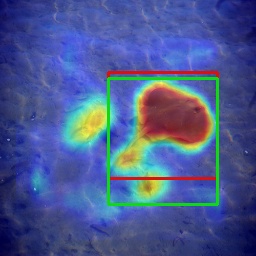}}
\end{minipage}
\begin{minipage}{0.11\linewidth}
\centerline{\includegraphics[scale=0.20]{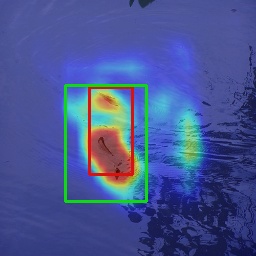}}
\end{minipage}
\begin{minipage}{0.11\linewidth}
\centerline{\includegraphics[scale=0.20]{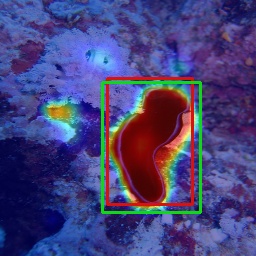}}
\end{minipage}
\begin{minipage}{0.11\linewidth}
\centerline{\includegraphics[scale=0.20]{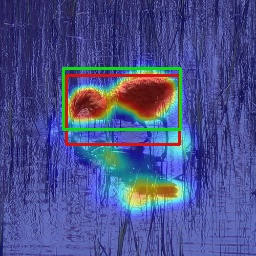}}
\end{minipage}
\begin{minipage}{0.11\linewidth}
\centerline{\includegraphics[scale=0.20]{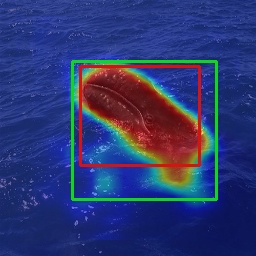}}
\end{minipage}
\begin{minipage}{0.11\linewidth}
\centerline{\includegraphics[scale=0.20]{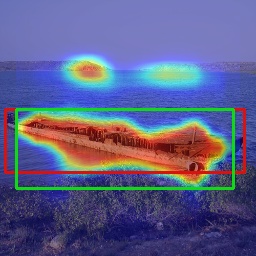}}
\end{minipage}
\\

\vspace{0.1cm}
\begin{minipage}{0.03\linewidth}
\centerline{(d)}
\end{minipage}
\begin{minipage}{0.11\linewidth}
\centerline{\includegraphics[scale=0.20]{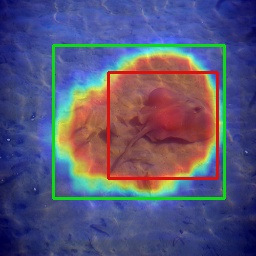}}
\end{minipage}
\begin{minipage}{0.11\linewidth}
\centerline{\includegraphics[scale=0.20]{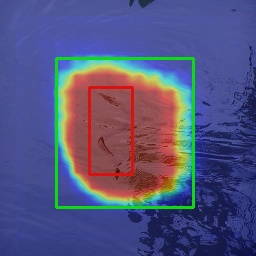}}
\end{minipage}
\begin{minipage}{0.11\linewidth}
\centerline{\includegraphics[scale=0.20]{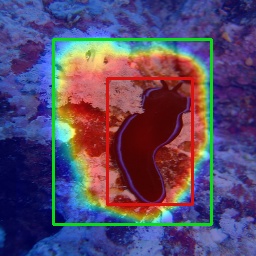}}
\end{minipage}
\begin{minipage}{0.11\linewidth}
\centerline{\includegraphics[scale=0.20]{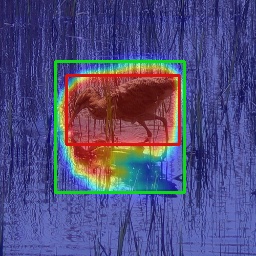}}
\end{minipage}
\begin{minipage}{0.11\linewidth}
\centerline{\includegraphics[scale=0.20]{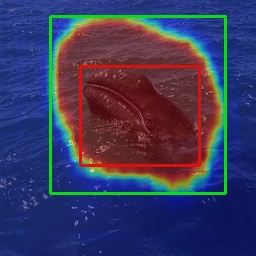}}
\end{minipage}
\begin{minipage}{0.11\linewidth}
\centerline{\includegraphics[scale=0.20]{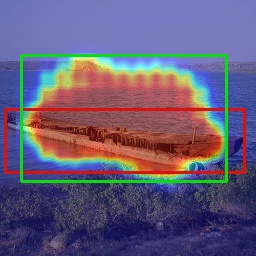}}
\end{minipage}
\\

\vspace{0.1cm}
\begin{minipage}{0.03\linewidth}
\centerline{(e)}
\end{minipage}
\begin{minipage}{0.11\linewidth}
\centerline{\includegraphics[scale=0.20]{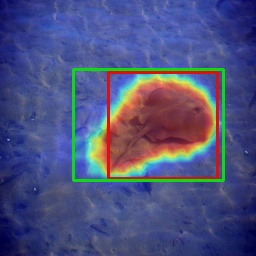}}
\end{minipage}
\begin{minipage}{0.11\linewidth}
\centerline{\includegraphics[scale=0.20]{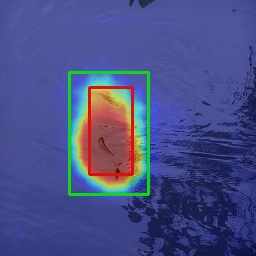}}
\end{minipage}
\begin{minipage}{0.11\linewidth}
\centerline{\includegraphics[scale=0.20]{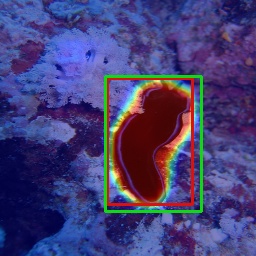}}
\end{minipage}
\begin{minipage}{0.11\linewidth}
\centerline{\includegraphics[scale=0.20]{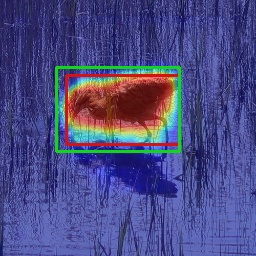}}
\end{minipage}
\begin{minipage}{0.11\linewidth}
\centerline{\includegraphics[scale=0.20]{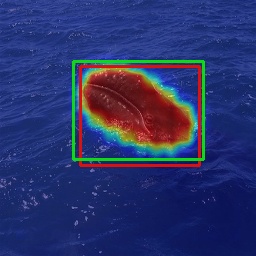}}
\end{minipage}
\begin{minipage}{0.11\linewidth}
\centerline{\includegraphics[scale=0.20]{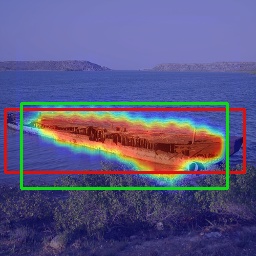}}
\end{minipage}
\caption{Qualitative comparison of the proposed method to the previous state-of-the-art method on the ILSVRC-2012 dataset~\citep{ILSVRC15}. (a) Image; (b) BAS~\citep{Wu_2022_CVPR} with MobileNetV1~\citep{mobilenetv1}; (c) Proposed method with MobileNetV1~\citep{mobilenetv1}; (d) BAS~\citep{Wu_2022_CVPR} with InceptionV3~\citep{inceptionv3}; (e) Proposed method with InceptionV3~\citep{inceptionv3}.}
\label{fig:result_ilsvrc}
\end{center}
\end{figure*}

\begin{figure*}[!t] 
\begin{center}
\begin{minipage}{0.03\linewidth}
\centerline{(a)}
\end{minipage}
\begin{minipage}{0.11\linewidth}
\centerline{\includegraphics[scale=0.20]{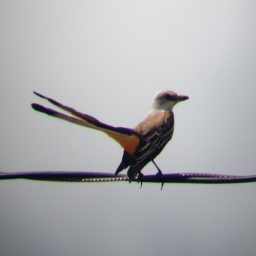}}
\end{minipage}
\begin{minipage}{0.11\linewidth}
\centerline{\includegraphics[scale=0.20]{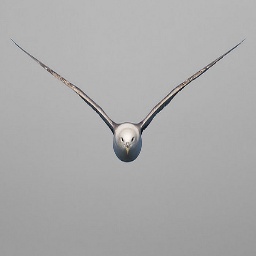}}
\end{minipage}
\begin{minipage}{0.11\linewidth}
\centerline{\includegraphics[scale=0.20]{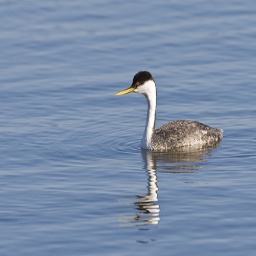}}
\end{minipage}
\begin{minipage}{0.11\linewidth}
\centerline{\includegraphics[scale=0.20]{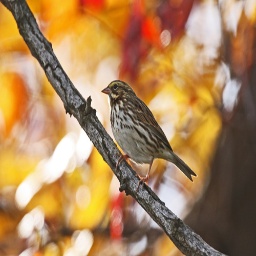}}
\end{minipage}
\begin{minipage}{0.11\linewidth}
\centerline{\includegraphics[scale=0.20]{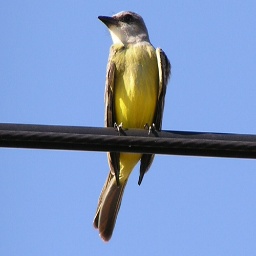}}
\end{minipage}
\begin{minipage}{0.11\linewidth}
\centerline{\includegraphics[scale=0.20]{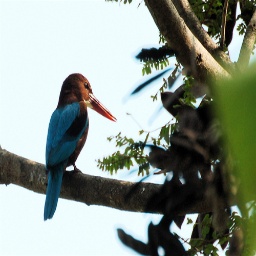}}
\end{minipage}
\\

\vspace{0.1cm}
\begin{minipage}{0.03\linewidth}
\centerline{(b)}
\end{minipage}
\begin{minipage}{0.11\linewidth}
\centerline{\includegraphics[scale=0.20]{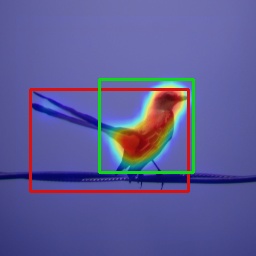}}
\end{minipage}
\begin{minipage}{0.11\linewidth}
\centerline{\includegraphics[scale=0.20]{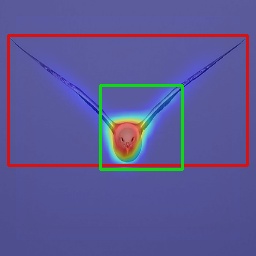}}
\end{minipage}
\begin{minipage}{0.11\linewidth}
\centerline{\includegraphics[scale=0.20]{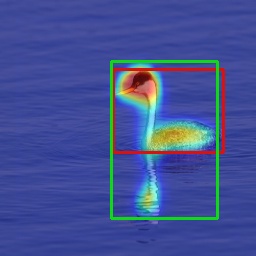}}
\end{minipage}
\begin{minipage}{0.11\linewidth}
\centerline{\includegraphics[scale=0.20]{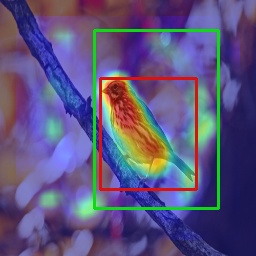}}
\end{minipage}
\begin{minipage}{0.11\linewidth}
\centerline{\includegraphics[scale=0.20]{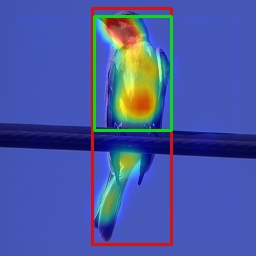}}
\end{minipage}
\begin{minipage}{0.11\linewidth}
\centerline{\includegraphics[scale=0.20]{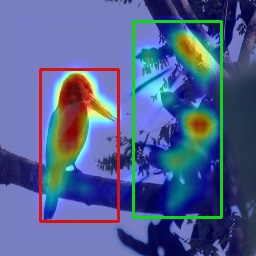}}
\end{minipage}
\\

\vspace{0.1cm}
\begin{minipage}{0.03\linewidth}
\centerline{(c)}
\end{minipage}
\begin{minipage}{0.11\linewidth}
\centerline{\includegraphics[scale=0.20]{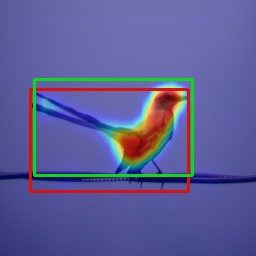}}
\end{minipage}
\begin{minipage}{0.11\linewidth}
\centerline{\includegraphics[scale=0.20]{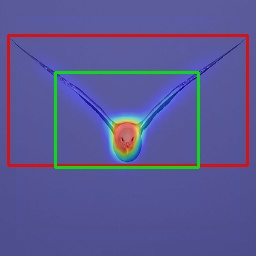}}
\end{minipage}
\begin{minipage}{0.11\linewidth}
\centerline{\includegraphics[scale=0.20]{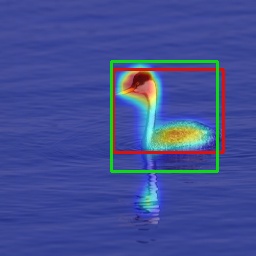}}
\end{minipage}
\begin{minipage}{0.11\linewidth}
\centerline{\includegraphics[scale=0.20]{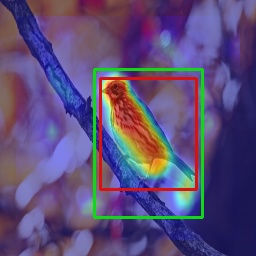}}
\end{minipage}
\begin{minipage}{0.11\linewidth}
\centerline{\includegraphics[scale=0.20]{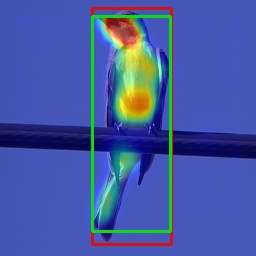}}
\end{minipage}
\begin{minipage}{0.11\linewidth}
\centerline{\includegraphics[scale=0.20]{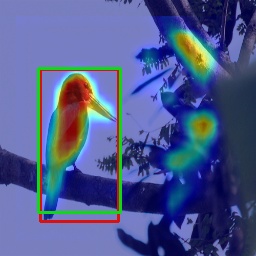}}
\end{minipage}
\\

\vspace{0.1cm}
\begin{minipage}{0.03\linewidth}
\centerline{(d)}
\end{minipage}
\begin{minipage}{0.11\linewidth}
\centerline{\includegraphics[scale=0.20]{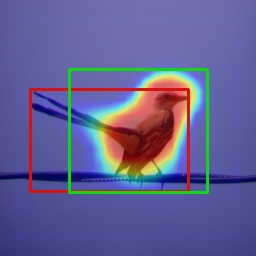}}
\end{minipage}
\begin{minipage}{0.11\linewidth}
\centerline{\includegraphics[scale=0.20]{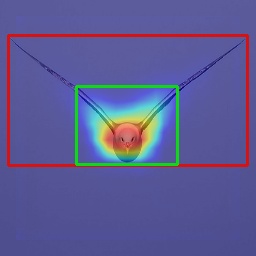}}
\end{minipage}
\begin{minipage}{0.11\linewidth}
\centerline{\includegraphics[scale=0.20]{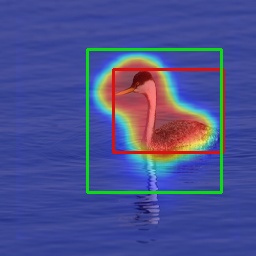}}
\end{minipage}
\begin{minipage}{0.11\linewidth}
\centerline{\includegraphics[scale=0.20]{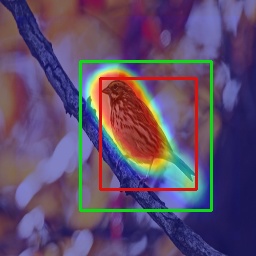}}
\end{minipage}
\begin{minipage}{0.11\linewidth}
\centerline{\includegraphics[scale=0.20]{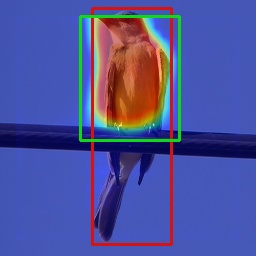}}
\end{minipage}
\begin{minipage}{0.11\linewidth}
\centerline{\includegraphics[scale=0.20]{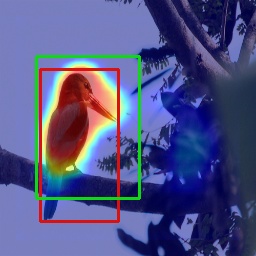}}
\end{minipage}
\\

\vspace{0.1cm}
\begin{minipage}{0.03\linewidth}
\centerline{(e)}
\end{minipage}
\begin{minipage}{0.11\linewidth}
\centerline{\includegraphics[scale=0.20]{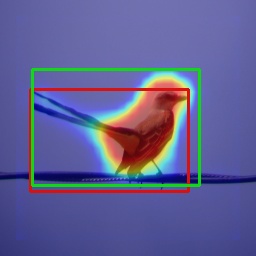}}
\end{minipage}
\begin{minipage}{0.11\linewidth}
\centerline{\includegraphics[scale=0.20]{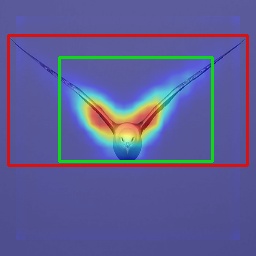}}
\end{minipage}
\begin{minipage}{0.11\linewidth}
\centerline{\includegraphics[scale=0.20]{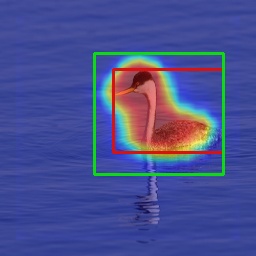}}
\end{minipage}
\begin{minipage}{0.11\linewidth}
\centerline{\includegraphics[scale=0.20]{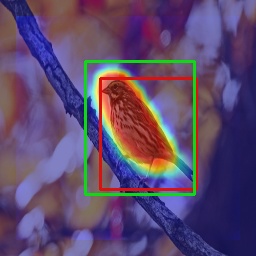}}
\end{minipage}
\begin{minipage}{0.11\linewidth}
\centerline{\includegraphics[scale=0.20]{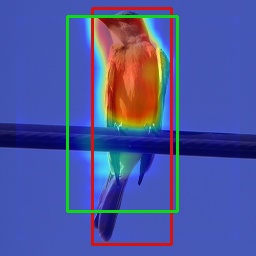}}
\end{minipage}
\begin{minipage}{0.11\linewidth}
\centerline{\includegraphics[scale=0.20]{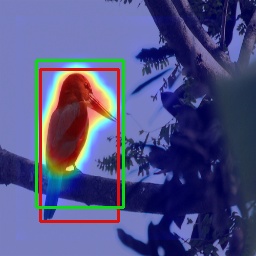}}
\end{minipage}
\caption{Qualitative comparison of the proposed method to the previous state-of-the-art method on the CUB-200-2011 dataset~\citep{cub_dataset}. (a) Image; (b) BAS~\citep{Wu_2022_CVPR} with MobileNetV1~\citep{mobilenetv1}; (c) Proposed method with MobileNetV1~\citep{mobilenetv1}; (d) BAS~\citep{Wu_2022_CVPR} with InceptionV3~\citep{inceptionv3}; (e) Proposed method with InceptionV3~\citep{inceptionv3}.}
\label{fig:result_cub}
\end{center}
\end{figure*}

\begin{figure*}[!t] 
\begin{center}
\begin{minipage}{0.03\linewidth}
\centerline{(a)}
\end{minipage}
\begin{minipage}{0.11\linewidth}
\centerline{\includegraphics[scale=0.20]{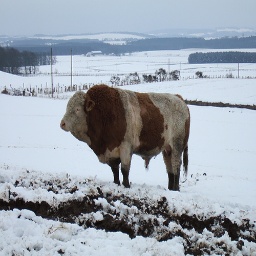}}
\end{minipage}
\begin{minipage}{0.11\linewidth}
\centerline{\includegraphics[scale=0.20]{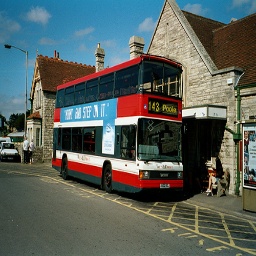}}
\end{minipage}
\begin{minipage}{0.11\linewidth}
\centerline{\includegraphics[scale=0.20]{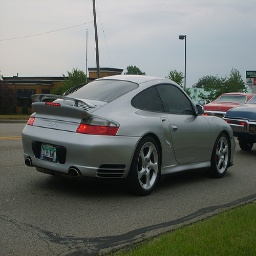}}
\end{minipage}
\begin{minipage}{0.11\linewidth}
\centerline{\includegraphics[scale=0.20]{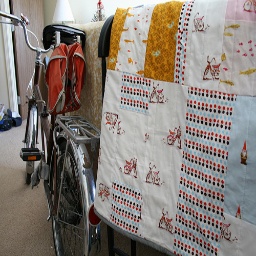}}
\end{minipage}
\begin{minipage}{0.11\linewidth}
\centerline{\includegraphics[scale=0.20]{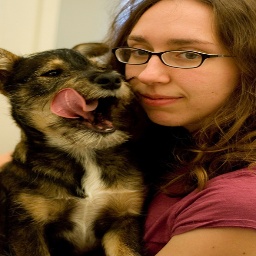}}
\end{minipage}
\begin{minipage}{0.11\linewidth}
\centerline{\includegraphics[scale=0.20]{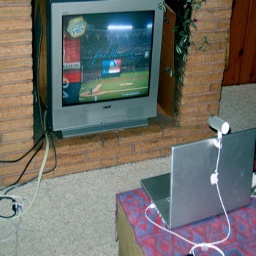}}
\end{minipage}
\\

\vspace{0.1cm}
\begin{minipage}{0.03\linewidth}
\centerline{(b)}
\end{minipage}
\begin{minipage}{0.11\linewidth}
\centerline{\includegraphics[scale=0.20]{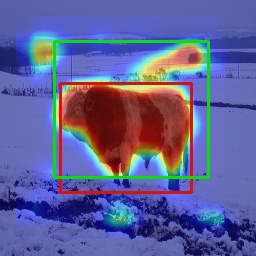}}
\end{minipage}
\begin{minipage}{0.11\linewidth}
\centerline{\includegraphics[scale=0.20]{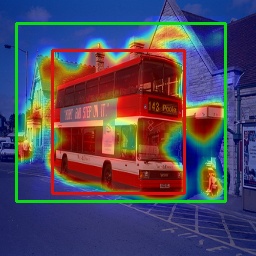}}
\end{minipage}
\begin{minipage}{0.11\linewidth}
\centerline{\includegraphics[scale=0.20]{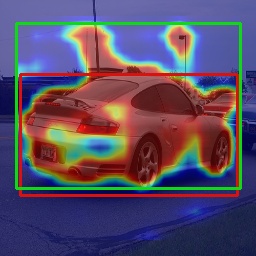}}
\end{minipage}
\begin{minipage}{0.11\linewidth}
\centerline{\includegraphics[scale=0.20]{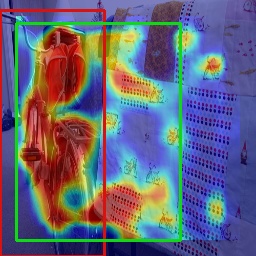}}
\end{minipage}
\begin{minipage}{0.11\linewidth}
\centerline{\includegraphics[scale=0.20]{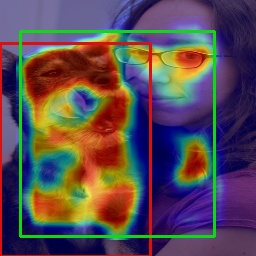}}
\end{minipage}
\begin{minipage}{0.11\linewidth}
\centerline{\includegraphics[scale=0.20]{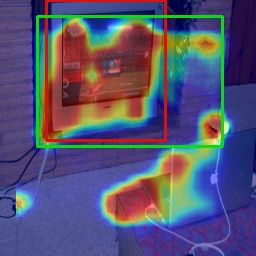}}
\end{minipage}
\\

\vspace{0.1cm}
\begin{minipage}{0.03\linewidth}
\centerline{(c)}
\end{minipage}
\begin{minipage}{0.11\linewidth}
\centerline{\includegraphics[scale=0.20]{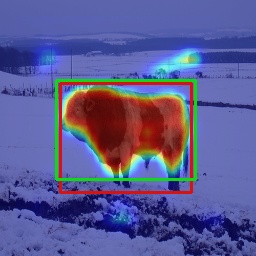}}
\end{minipage}
\begin{minipage}{0.11\linewidth}
\centerline{\includegraphics[scale=0.20]{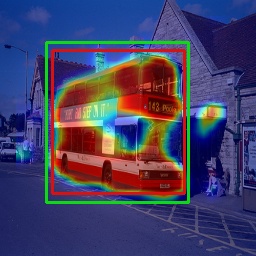}}
\end{minipage}
\begin{minipage}{0.11\linewidth}
\centerline{\includegraphics[scale=0.20]{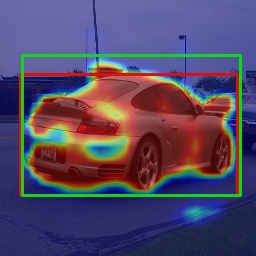}}
\end{minipage}
\begin{minipage}{0.11\linewidth}
\centerline{\includegraphics[scale=0.20]{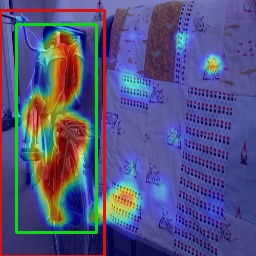}}
\end{minipage}
\begin{minipage}{0.11\linewidth}
\centerline{\includegraphics[scale=0.20]{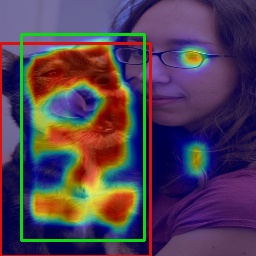}}
\end{minipage}
\begin{minipage}{0.11\linewidth}
\centerline{\includegraphics[scale=0.20]{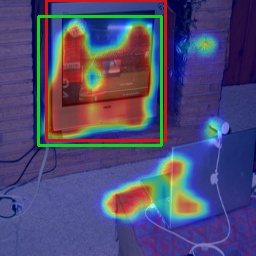}}
\end{minipage}
\caption{Qualitative comparison of the proposed method with the previous state-of-the-art method on the PASCAL VOC 2012 classification dataset~\citep{pascalvoc}. (a) Image; (b) BAS~\citep{Wu_2022_CVPR} with MobileNetV1~\citep{mobilenetv1}; (c) Proposed method with MobileNetV1~\citep{mobilenetv1}.}
\label{fig:result_voc}
\end{center}
\end{figure*}

Figures~\ref{fig:result_ilsvrc},~\ref{fig:result_cub}, and~\ref{fig:result_voc} show the qualitative results for localization on the ILSVRC-2012 dataset, the CUB-200-2011 dataset, and the PASCAL VOC 2012 classification dataset, respectively. The results correspond to the experiment setting of GT-known. The red and green bounding boxes represent the ground-truth labels and the results of the corresponding methods, respectively. Each row shows the input image, the result of BAS~\citep{Wu_2022_CVPR} with MobileNetV1, the result of the proposed method with MobileNetV1, the result of BAS~\citep{Wu_2022_CVPR} with InceptionV3, and the result of the proposed method with InceptionV3. The visualized results demonstrate that the proposed method effectively suppresses background areas while activating the entire object regions.

\subsection{Analysis}
\tref{tab:result_ablation1} shows the effects of utilizing additional loss terms by experimenting with the MobileNetV1 backbone and the CUB-200-2011 dataset. The baseline method uses the loss terms in~\eref{eq:total_loss} excluding $\calL_{pseudo}$, $\calL_{ae\text{-}fg}$, and $\calL_{ae}$. By also using $\calL_{pseudo}$, the accuracy is improved by absolute 0.64\%, 0.07\%, and 0.13\% in Top-1, Top-5, and GT-known localization, respectively. $\calL_{ae\text{-}fg}$ improves the performances further by 0.05\%, 0.64\%, and 0.35\%. Finally, by $\calL_{ae}$, the performance is improved by 0.82\%, 0.29\%, and 0.20\%.

\begin{table*}[!t]
\centering
\small
\begin{minipage}{0.95\linewidth}
\caption{Ablation study on loss terms using the MobileNetV1 backbone~\citep{mobilenetv1} and the CUB-200-2011 dataset~\citep{cub_dataset}.}
\label{tab:result_ablation1}
\renewcommand{\arraystretch}{1.} 
\centering
\begin{tabu}{X[c,m]|X[c,m]|X[c,m]|X[c,m]|X[c,m]|X[c,m]|X[c,m]|X[c,m]} 
\hline
\multicolumn{4}{c|}{Method} & \multirow{2}{*}{Top-1} & \multirow{2}{*}{Top-5} & \multirow{2}{*}{GT-known} & \multirow{2}{*}{mIoU} \\ 
\tabucline{1-4}
Baseline & $\calL_{pseudo}$ & $\calL_{ae\text{-}fg}$ & $\calL_{ae}$ & & & \\ 
\hline\hline
\checkmark & $-$ & $-$ & $-$ & 69.77 & 86.00 & 92.35 & 54.51 \\
\checkmark & \checkmark & $-$ & $-$ & 70.41 & 86.07 & 92.48 & 55.22 \\
\checkmark & \checkmark & \checkmark & $-$ & 70.46 & 86.71 & 92.83 & 55.32 \\
\checkmark & \checkmark & \checkmark & \checkmark & \textbf{71.28}& \textbf{87.00}&\textbf{93.03} & \textbf{56.08} \\
\hline  		
\end{tabu}
\end{minipage}
\end{table*}

\tref{tab:result_ablation2} shows an ablation study on the weighting coefficients in the total loss. The ablation study is conducted by using the MobileNetV1 backbone~\citep{mobilenetv1} and the CUB-200-2011 dataset~\citep{cub_dataset}. The results of the top three rows correspond to them in~\tref{tab:result_ablation1} since excluding a loss term is equivalent to making the corresponding coefficient zero. All the results in this table are better than the previous state-of-the-art method. This ablation study demonstrates that the proposed method outperforms the previous state-of-the-art method even with varying hyperparameters.

\begin{table*}[!t]
\centering
\small
\begin{minipage}{0.95\linewidth}
\caption{Ablation study of weighting coefficients in the total loss by using the MobileNetV1 backbone~\citep{mobilenetv1} and the CUB-200-2011 dataset~\citep{cub_dataset}.}
\label{tab:result_ablation2}
\renewcommand{\arraystretch}{1.} 
\centering
\begin{tabular}{ *{6}{>{\centering}m{0.06\textwidth}|} *{3}{>{\centering}m{0.095\textwidth}|} >{\centering\arraybackslash}m{0.095\textwidth} } 
\hline
\multicolumn{6}{c|}{Weighting coefficients} & \multirow{2}{*}{Top-1} & \multirow{2}{*}{Top-5} & \multirow{2}{*}{GT-known} & \multirow{2}{*}{mIoU} \\ 
\cline{1-6}
$\gamma_1$ & $\gamma_2$ & $\gamma_3$ & $\gamma_4$ & $\gamma_5$ & $\gamma_6$ & & & & \\ 
\hline\hline
0.5 & 0 & 0   & 0   & 1.0 & 1.5  & 69.77 & 86.00 & 92.35 & 54.51 \\
0.5 & 0 & 0   & 0.1 & 1.0 & 1.5  & 70.41 & 86.07 & 92.48 & 55.22 \\
0.5 & 0 & 0.1 & 0.1 & 1.0 & 1.5  & 70.46 & 86.71 & 92.83 & 55.32  \\
0.5 & 0.6 & 0.1 & 0.1 & 1.0 & 1.5 & 71.63 & 86.54 & 92.68 & 56.39 \\
0.5 & 0.5 & 0.2 & 0.1 & 1.0 & 1.5 & 70.96 & 86.62 & 92.48 & 55.97 \\
0.5 & 0.5 & 0.1 & 0.12 & 1.0 & 1.5 & 71.24 & 86.32 & 92.55 & 55.92 \\
0.5 & 0.5 & 0.1 & 0.1 & 1.1 & 1.5 & 71.21 & 86.96 & 92.49 & 56.39 \\
0.5 & 0.5 & 0.1 & 0.1 & 1.0 & 1.6 & 70.82 & 86.47 & 92.41 & 55.82 \\
0.5 & 0.5 & 0.1 & 0.1 & 1.0 & 1.5 & \textbf{71.28}& \textbf{87.00}&\textbf{93.03} & \textbf{56.08}  \\
\hline  		
\end{tabular}
\end{minipage}
\end{table*}

\tref{tab:ablation_threshold} shows the analysis of varying hyperparameters in adversarial erasing and pseudo label generation. This analysis was conducted using the MobileNetV1 backbone~\citep{mobilenetv1} and the CUB-200-2011 dataset~\citep{cub_dataset}. Experimental results demonstrate that the proposed method consistently outperforms the previous state-of-the-art method. As the hyperparameters on the bottom row achieves the highest accuracy for three metrics among four, the parameters $t_1$, $t_2$, $t_3$, and $t_4$ are set as 0.8, 0.8, 0.4, and 0.1, respectively. The degrade of accuracy when $t_4$ is reduced to 0.5, is natural because $t_4=0$ corresponds to not generating and utilizing pseudo labels of zeros.

\begin{table}[!t]
\centering
\small
\begin{minipage}{0.95\linewidth}
\caption{Analysis of parameters for adversarial erasing and pseudo label generation using the MobileNetV1 backbone~\citep{mobilenetv1} and the CUB-200-2011 dataset~\citep{cub_dataset}.}
\label{tab:ablation_threshold}
\renewcommand{\arraystretch}{1.} 
\centering
\begin{tabular}{ *{4}{>{\centering}m{0.055\textwidth}|} *{3}{>{\centering}m{0.09\textwidth}|} >{\centering\arraybackslash}m{0.09\textwidth} } 
\hline
\multicolumn{4}{c|}{Parameter} & \multirow{2}{*}{Top-1} & \multirow{2}{*}{Top-5} & GT- & \multirow{2}{*}{mIoU} \\ 
\cline{1-4}
$t_1$ & $t_2$ & $t_3$ & $t_4$ & & & known & \\ 
\hline\hline
0.75 & 0.8 & 0.4 & 0.1 & 71.13 & 86.52 & 92.41 & 56.12 \\
0.85 & 0.8 & 0.4 & 0.1 & 70.66	& 86.43 & 92.56 & 55.48 \\
0.8 & 0.75 & 0.4 & 0.1 & 71.09	& 86.72 & 92.84 & 55.72 \\
0.8 & 0.85 & 0.4 & 0.1 & 71.29	& 86.12	& 92.49 & 55.98 \\
0.8 & 0.8 & 0.35 & 0.1 & 71.11	& 86.68	& 92.88 & 55.77 \\
0.8 & 0.8 & 0.45 & 0.1 & 71.26 & 86.75	& 92.63 & \textbf{56.13} \\
0.8 & 0.8 & 0.4 & 0.05 & 71.03	& 86.51	& 92.32 & 55.99 \\
0.8 & 0.8 & 0.4 & 0.15 & 71.10	& 86.41	& 92.49 & 56.12 \\
0.8 & 0.8 & 0.4 & 0.1 & \textbf{71.28}& \textbf{87.00}&\textbf{93.03} & 56.08  \\
\hline  		
\end{tabular}
\end{minipage}
\end{table}

\tref{tab:ablation_resolution} shows the analysis of using different spatial resolutions for the shared feature map $\mF^f$. The experiments are conducted using the MobileNetV1 backbone~\citep{mobilenetv1} and the CUB-200-2011 dataset~\citep{cub_dataset}. The backbone consists of one standard convolution layer and 11 depthwise separable convolutions, following the original MobileNetV1~\citep{mobilenetv1}. The resolution of $14 \times 14$ corresponds to the case using the original MobileNetV1 backbone~\citep{mobilenetv1}. The resolution of $28 \times 28$ is achieved by replacing the last depthwise separable convolution layer in the backbone, which had a stride of two, with a layer that has a stride of one. The experimental results demonstrate that replacing the last layer with a stride of two with a stride of one improves localization accuracies.

\begin{table}[!t]
\centering
\small
\begin{minipage}{0.95\linewidth}
\caption{Analysis of spatial resolution of the shared feature map $\mF^f$ using the MobileNetV1 backbone~\citep{mobilenetv1} and the CUB-200-2011 dataset~\citep{cub_dataset}.}
\label{tab:ablation_resolution}
\renewcommand{\arraystretch}{1.} 
\centering
\begin{tabular}{ >{\centering}m{0.2\textwidth}| >{\centering}m{0.14\textwidth}| >{\centering}m{0.14\textwidth}| >{\centering}m{0.14\textwidth}| >{\centering\arraybackslash}m{0.14\textwidth} } 
\hline
Resolution & Top-1 & Top-5 & GT-known & mIoU \\ 
\hline\hline
$ 14 \times 14 $ & 67.40	& 84.53	& 92.13	& 49.79 \\
$ 28 \times 28 $ & \textbf{71.28}& \textbf{87.00}&\textbf{93.03} & \textbf{56.08}  \\
\hline  		
\end{tabular}
\end{minipage}
\end{table}

\section{CONCLUSION}
We presented a framework that consists of a shared feature extractor, a classifier, and a localizer for weakly-supervised object localization. Since the localizer is trained using only image-level class labels instead of detailed location information, it often localizes only the most discriminative region rather than the entire object region. Therefore, to improve localization accuracy by localizing the entire region rather than just the discriminative part, we proposed two loss terms using adversarial erasing. One is computed using the adversarially erased feature map, and the other is calculated using the features from the adversarially erased foreground region. We experimentally demonstrated that the two losses complement each other. Additionally, we proposed a loss term based on pixel-level pseudo labels. Due to the lack of detailed location information, we generated pixel-level pseudo labels and used them to guide the localizer to suppress activation values in the background and increase them in the foreground. We utilized the proposed method to train the networks with two different backbones and on three publicly available datasets. The experimental results confirm that the proposed method outperforms the previous state-of-the-art methods in all evaluated metrics.

Regarding the limitations of this work, we utilized image classification datasets, which usually consist of only a few objects per image. This might not accurately represent real-world scenarios where images can contain a higher number of objects. Furthermore, we assigned a single image-level class label to each image, whereas images with multiple objects could potentially be labeled with multiple class labels. Addressing these limitations could serve as a valuable direction for future research.

\section*{Acknowledgments}
This work was supported by the Institute of Information \& communications Technology Planning \& Evaluation (IITP) grant funded by the Korea government(MSIT) (No.2020-0-00994, Development of autonomous VR and AR content generation technology reflecting usage environment). 

\vspace{0.5cm}

\bibliographystyle{elsarticle-harv}
\bibliography{mybibfile}

\begin{thebibliography}{46}
\expandafter\ifx\csname natexlab\endcsname\relax\def\natexlab#1{#1}\fi
\providecommand{\url}[1]{\texttt{#1}}
\providecommand{\href}[2]{#2}
\providecommand{\path}[1]{#1}
\providecommand{\DOIprefix}{doi:}
\providecommand{\ArXivprefix}{arXiv:}
\providecommand{\URLprefix}{URL: }
\providecommand{\Pubmedprefix}{pmid:}
\providecommand{\doi}[1]{\href{http://dx.doi.org/#1}{\path{#1}}}
\providecommand{\Pubmed}[1]{\href{pmid:#1}{\path{#1}}}
\providecommand{\bibinfo}[2]{#2}
\ifx\xfnm\relax \def\xfnm[#1]{\unskip,\space#1}\fi
\bibitem[{Afaq and Manocha(2022)}]{Afaq2022_r3}
\bibinfo{author}{Afaq, Y.}, \bibinfo{author}{Manocha, A.},
  \bibinfo{year}{2022}.
\newblock \bibinfo{title}{Multi-resolution-based deep learning approach for
  rice field monitoring}.
\newblock \bibinfo{journal}{Canadian Journal of Remote Sensing}
  \bibinfo{volume}{48}, \bibinfo{pages}{278--298}.
\newblock \DOIprefix\doi{10.1080/07038992.2021.2010036}.
\bibitem[{Bae et~al.(2020)Bae, Noh and Kim}]{Bae2020}
\bibinfo{author}{Bae, W.}, \bibinfo{author}{Noh, J.}, \bibinfo{author}{Kim,
  G.}, \bibinfo{year}{2020}.
\newblock \bibinfo{title}{Rethinking class activation mapping for weakly
  supervised object localization}, in: \bibinfo{editor}{Vedaldi, A.},
  \bibinfo{editor}{Bischof, H.}, \bibinfo{editor}{Brox, T.},
  \bibinfo{editor}{Frahm, J.M.} (Eds.), \bibinfo{booktitle}{Computer Vision --
  ECCV 2020}, \bibinfo{publisher}{Springer International Publishing},
  \bibinfo{address}{Cham}. pp. \bibinfo{pages}{618--634}.
\bibitem[{Cai et~al.(2023)Cai, Xiao, Zeng, Gong and Ni}]{Cai2023_EAAI}
\bibinfo{author}{Cai, X.}, \bibinfo{author}{Xiao, R.}, \bibinfo{author}{Zeng,
  Z.}, \bibinfo{author}{Gong, P.}, \bibinfo{author}{Ni, Y.},
  \bibinfo{year}{2023}.
\newblock \bibinfo{title}{Itran: A novel transformer-based approach for
  industrial anomaly detection and localization}.
\newblock \bibinfo{journal}{Engineering Applications of Artificial
  Intelligence} \bibinfo{volume}{125}, \bibinfo{pages}{106677}.
\newblock \URLprefix
  \url{https://www.sciencedirect.com/science/article/pii/S0952197623008618},
  \DOIprefix\doi{https://doi.org/10.1016/j.engappai.2023.106677}.
\bibitem[{Chen and Liu(2018)}]{Chen2018_BLS}
\bibinfo{author}{Chen, C.L.P.}, \bibinfo{author}{Liu, Z.},
  \bibinfo{year}{2018}.
\newblock \bibinfo{title}{Broad learning system: An effective and efficient
  incremental learning system without the need for deep architecture}.
\newblock \bibinfo{journal}{IEEE Transactions on Neural Networks and Learning
  Systems} \bibinfo{volume}{29}, \bibinfo{pages}{10--24}.
\newblock \DOIprefix\doi{10.1109/TNNLS.2017.2716952}.
\bibitem[{Choe et~al.(2021)Choe, Lee and Shim}]{Choe2021}
\bibinfo{author}{Choe, J.}, \bibinfo{author}{Lee, S.}, \bibinfo{author}{Shim,
  H.}, \bibinfo{year}{2021}.
\newblock \bibinfo{title}{Attention-based dropout layer for weakly supervised
  single object localization and semantic segmentation}.
\newblock \bibinfo{journal}{IEEE Transactions on Pattern Analysis and Machine
  Intelligence} \bibinfo{volume}{43}, \bibinfo{pages}{4256--4271}.
\newblock \DOIprefix\doi{10.1109/TPAMI.2020.2999099}.
\bibitem[{Choe and Shim(2019)}]{Choe2019}
\bibinfo{author}{Choe, J.}, \bibinfo{author}{Shim, H.}, \bibinfo{year}{2019}.
\newblock \bibinfo{title}{Attention-based dropout layer for weakly supervised
  object localization}, in: \bibinfo{booktitle}{2019 IEEE/CVF Conference on
  Computer Vision and Pattern Recognition (CVPR)}, pp.
  \bibinfo{pages}{2214--2223}.
\newblock \DOIprefix\doi{10.1109/CVPR.2019.00232}.
\bibitem[{Dosovitskiy et~al.(2020)Dosovitskiy, Beyer, Kolesnikov, Weissenborn,
  Zhai, Unterthiner, Dehghani, Minderer, Heigold, Gelly, Uszkoreit and
  Houlsby}]{visionTransformer2020}
\bibinfo{author}{Dosovitskiy, A.}, \bibinfo{author}{Beyer, L.},
  \bibinfo{author}{Kolesnikov, A.}, \bibinfo{author}{Weissenborn, D.},
  \bibinfo{author}{Zhai, X.}, \bibinfo{author}{Unterthiner, T.},
  \bibinfo{author}{Dehghani, M.}, \bibinfo{author}{Minderer, M.},
  \bibinfo{author}{Heigold, G.}, \bibinfo{author}{Gelly, S.},
  \bibinfo{author}{Uszkoreit, J.}, \bibinfo{author}{Houlsby, N.},
  \bibinfo{year}{2020}.
\newblock \bibinfo{title}{An image is worth 16x16 words: Transformers for image
  recognition at scale}.
\newblock \bibinfo{journal}{CoRR} \bibinfo{volume}{abs/2010.11929}.
\newblock \URLprefix \url{https://arxiv.org/abs/2010.11929},
  \href{http://arxiv.org/abs/2010.11929}{{\tt arXiv:2010.11929}}.
\bibitem[{Everingham et~al.(2010)Everingham, Van~Gool, Williams, Winn and
  Zisserman}]{pascalvoc}
\bibinfo{author}{Everingham, M.}, \bibinfo{author}{Van~Gool, L.},
  \bibinfo{author}{Williams, C.K.I.}, \bibinfo{author}{Winn, J.},
  \bibinfo{author}{Zisserman, A.}, \bibinfo{year}{2010}.
\newblock \bibinfo{title}{The pascal visual object classes (voc) challenge}.
\newblock \bibinfo{journal}{International Journal of Computer Vision}
  \bibinfo{volume}{88}, \bibinfo{pages}{303--338}.
\newblock \URLprefix \url{https://doi.org/10.1007/s11263-009-0275-4},
  \DOIprefix\doi{10.1007/s11263-009-0275-4}.
\bibitem[{Gao et~al.(2023)Gao, Guo, Huang and Chen}]{Gao2023_r2}
\bibinfo{author}{Gao, S.}, \bibinfo{author}{Guo, G.}, \bibinfo{author}{Huang,
  H.}, \bibinfo{author}{Chen, C.L.P.}, \bibinfo{year}{2023}.
\newblock \bibinfo{title}{Go deep or broad? exploit hybrid network architecture
  for weakly supervised object classification and localization}.
\newblock \bibinfo{journal}{IEEE Transactions on Neural Networks and Learning
  Systems} , \bibinfo{pages}{1--14}\DOIprefix\doi{10.1109/TNNLS.2022.3225180}.
\bibitem[{Gao et~al.(2021)Gao, Wan, Pan, Peng, Tian, Han, Zhou and
  Ye}]{Gao2021}
\bibinfo{author}{Gao, W.}, \bibinfo{author}{Wan, F.}, \bibinfo{author}{Pan,
  X.}, \bibinfo{author}{Peng, Z.}, \bibinfo{author}{Tian, Q.},
  \bibinfo{author}{Han, Z.}, \bibinfo{author}{Zhou, B.}, \bibinfo{author}{Ye,
  Q.}, \bibinfo{year}{2021}.
\newblock \bibinfo{title}{Ts-cam: Token semantic coupled attention map for
  weakly supervised object localization}, in: \bibinfo{booktitle}{2021 IEEE/CVF
  International Conference on Computer Vision (ICCV)}, pp.
  \bibinfo{pages}{2866--2875}.
\newblock \DOIprefix\doi{10.1109/ICCV48922.2021.00288}.
\bibitem[{Guo et~al.(2021)Guo, Han, Wan and Zhang}]{Guo2021}
\bibinfo{author}{Guo, G.}, \bibinfo{author}{Han, J.}, \bibinfo{author}{Wan,
  F.}, \bibinfo{author}{Zhang, D.}, \bibinfo{year}{2021}.
\newblock \bibinfo{title}{Strengthen learning tolerance for weakly supervised
  object localization}, in: \bibinfo{booktitle}{2021 IEEE/CVF Conference on
  Computer Vision and Pattern Recognition (CVPR)}, pp.
  \bibinfo{pages}{7399--7408}.
\newblock \DOIprefix\doi{10.1109/CVPR46437.2021.00732}.
\bibitem[{Gupta et~al.(2022)Gupta, Lakhotia, Rawat and
  Tallamraju}]{Gupta_2022_CVPR}
\bibinfo{author}{Gupta, S.}, \bibinfo{author}{Lakhotia, S.},
  \bibinfo{author}{Rawat, A.}, \bibinfo{author}{Tallamraju, R.},
  \bibinfo{year}{2022}.
\newblock \bibinfo{title}{Vitol: Vision transformer for weakly supervised
  object localization}, in: \bibinfo{booktitle}{Proceedings of the IEEE/CVF
  Conference on Computer Vision and Pattern Recognition (CVPR) Workshops}, pp.
  \bibinfo{pages}{4101--4110}.
\bibitem[{Howard et~al.(2017)Howard, Zhu, Chen, Kalenichenko, Wang, Weyand,
  Andreetto and Adam}]{mobilenetv1}
\bibinfo{author}{Howard, A.G.}, \bibinfo{author}{Zhu, M.},
  \bibinfo{author}{Chen, B.}, \bibinfo{author}{Kalenichenko, D.},
  \bibinfo{author}{Wang, W.}, \bibinfo{author}{Weyand, T.},
  \bibinfo{author}{Andreetto, M.}, \bibinfo{author}{Adam, H.},
  \bibinfo{year}{2017}.
\newblock \bibinfo{title}{Mobilenets: Efficient convolutional neural networks
  for mobile vision applications}.
\newblock \bibinfo{journal}{CoRR} \bibinfo{volume}{abs/1704.04861}.
\newblock \URLprefix \url{http://arxiv.org/abs/1704.04861},
  \href{http://arxiv.org/abs/1704.04861}{{\tt arXiv:1704.04861}}.
\bibitem[{Jang et~al.(2023)Jang, Kwon, Jin and Kim}]{jang2023Weakly}
\bibinfo{author}{Jang, S.}, \bibinfo{author}{Kwon, J.}, \bibinfo{author}{Jin,
  K.}, \bibinfo{author}{Kim, Y.}, \bibinfo{year}{2023}.
\newblock \bibinfo{title}{Weakly supervised semantic segmentation via graph
  recalibration with scaling weight unit}.
\newblock \bibinfo{journal}{Engineering Applications of Artificial
  Intelligence} \bibinfo{volume}{119}, \bibinfo{pages}{105706}.
\newblock \URLprefix
  \url{https://www.sciencedirect.com/science/article/pii/S0952197622006960},
  \DOIprefix\doi{https://doi.org/10.1016/j.engappai.2022.105706}.
\bibitem[{Jiang et~al.(2021)Jiang, Zhang, Hou, Cheng and Wei}]{Jiang2021}
\bibinfo{author}{Jiang, P.T.}, \bibinfo{author}{Zhang, C.B.},
  \bibinfo{author}{Hou, Q.}, \bibinfo{author}{Cheng, M.M.},
  \bibinfo{author}{Wei, Y.}, \bibinfo{year}{2021}.
\newblock \bibinfo{title}{Layercam: Exploring hierarchical class activation
  maps for localization}.
\newblock \bibinfo{journal}{IEEE Transactions on Image Processing}
  \bibinfo{volume}{30}, \bibinfo{pages}{5875--5888}.
\newblock \DOIprefix\doi{10.1109/TIP.2021.3089943}.
\bibitem[{Kim et~al.(2023)Kim, Cha and Kang}]{Kim2023Multiscale}
\bibinfo{author}{Kim, D.M.}, \bibinfo{author}{Cha, S.}, \bibinfo{author}{Kang,
  B.}, \bibinfo{year}{2023}.
\newblock \bibinfo{title}{Multiscale vision transformer with deep
  clustering-guided refinement for weakly supervised object localization}, in:
  \bibinfo{booktitle}{2023 IEEE International Conference on Visual
  Communications and Image Processing (VCIP)}, pp. \bibinfo{pages}{1--5}.
\newblock \DOIprefix\doi{10.1109/VCIP59821.2023.10402750}.
\bibitem[{Kim et~al.(2021a)Kim, Choe, Yun and Kwak}]{normalization2021}
\bibinfo{author}{Kim, J.}, \bibinfo{author}{Choe, J.}, \bibinfo{author}{Yun,
  S.}, \bibinfo{author}{Kwak, N.}, \bibinfo{year}{2021}a.
\newblock \bibinfo{title}{Normalization matters in weakly supervised object
  localization}, in: \bibinfo{booktitle}{2021 IEEE/CVF International Conference
  on Computer Vision (ICCV)}, pp. \bibinfo{pages}{3407--3416}.
\newblock \DOIprefix\doi{10.1109/ICCV48922.2021.00341}.
\bibitem[{Kim et~al.(2021b)Kim, Choe, Akata and Oh}]{Kim2021}
\bibinfo{author}{Kim, J.M.}, \bibinfo{author}{Choe, J.},
  \bibinfo{author}{Akata, Z.}, \bibinfo{author}{Oh, S.J.},
  \bibinfo{year}{2021}b.
\newblock \bibinfo{title}{Keep calm and improve visual feature attribution},
  in: \bibinfo{booktitle}{2021 IEEE/CVF International Conference on Computer
  Vision (ICCV)}, pp. \bibinfo{pages}{8330--8340}.
\newblock \DOIprefix\doi{10.1109/ICCV48922.2021.00824}.
\bibitem[{Li et~al.(2023)Li, Miao, Ma, Shuang and Huang}]{Li2023_EAAI}
\bibinfo{author}{Li, Y.}, \bibinfo{author}{Miao, N.}, \bibinfo{author}{Ma, L.},
  \bibinfo{author}{Shuang, F.}, \bibinfo{author}{Huang, X.},
  \bibinfo{year}{2023}.
\newblock \bibinfo{title}{Transformer for object detection: Review and
  benchmark}.
\newblock \bibinfo{journal}{Engineering Applications of Artificial
  Intelligence} \bibinfo{volume}{126}, \bibinfo{pages}{107021}.
\newblock \URLprefix
  \url{https://www.sciencedirect.com/science/article/pii/S0952197623012058},
  \DOIprefix\doi{https://doi.org/10.1016/j.engappai.2023.107021}.
\bibitem[{Liu et~al.(2023)Liu, Zhang, Wang, Yang, Qiu, Coleman and
  Kerr}]{Liu2023_EAAI}
\bibinfo{author}{Liu, Y.}, \bibinfo{author}{Zhang, Y.}, \bibinfo{author}{Wang,
  Z.}, \bibinfo{author}{Yang, F.}, \bibinfo{author}{Qiu, F.},
  \bibinfo{author}{Coleman, S.}, \bibinfo{author}{Kerr, D.},
  \bibinfo{year}{2023}.
\newblock \bibinfo{title}{A novel seminar learning framework for weakly
  supervised salient object detection}.
\newblock \bibinfo{journal}{Engineering Applications of Artificial
  Intelligence} \bibinfo{volume}{126}, \bibinfo{pages}{106961}.
\newblock \URLprefix
  \url{https://www.sciencedirect.com/science/article/pii/S0952197623011454},
  \DOIprefix\doi{https://doi.org/10.1016/j.engappai.2023.106961}.
\bibitem[{Lu et~al.(2020)Lu, Jia, Xie, Shen, Zhou and Duan}]{Lu2020}
\bibinfo{author}{Lu, W.}, \bibinfo{author}{Jia, X.}, \bibinfo{author}{Xie, W.},
  \bibinfo{author}{Shen, L.}, \bibinfo{author}{Zhou, Y.},
  \bibinfo{author}{Duan, J.}, \bibinfo{year}{2020}.
\newblock \bibinfo{title}{Geometry constrained weakly supervised object
  localization}, in: \bibinfo{editor}{Vedaldi, A.}, \bibinfo{editor}{Bischof,
  H.}, \bibinfo{editor}{Brox, T.}, \bibinfo{editor}{Frahm, J.M.} (Eds.),
  \bibinfo{booktitle}{Computer Vision -- ECCV 2020},
  \bibinfo{publisher}{Springer International Publishing},
  \bibinfo{address}{Cham}. pp. \bibinfo{pages}{481--496}.
\bibitem[{Meng et~al.(2021)Meng, Zhang, Tian, Zhang and Wu}]{Meng2021}
\bibinfo{author}{Meng, M.}, \bibinfo{author}{Zhang, T.}, \bibinfo{author}{Tian,
  Q.}, \bibinfo{author}{Zhang, Y.}, \bibinfo{author}{Wu, F.},
  \bibinfo{year}{2021}.
\newblock \bibinfo{title}{Foreground activation maps for weakly supervised
  object localization}, in: \bibinfo{booktitle}{2021 IEEE/CVF International
  Conference on Computer Vision (ICCV)}, pp. \bibinfo{pages}{3365--3375}.
\newblock \DOIprefix\doi{10.1109/ICCV48922.2021.00337}.
\bibitem[{Oquab et~al.(2015)Oquab, Bottou, Laptev and Sivic}]{Oquab2015}
\bibinfo{author}{Oquab, M.}, \bibinfo{author}{Bottou, L.},
  \bibinfo{author}{Laptev, I.}, \bibinfo{author}{Sivic, J.},
  \bibinfo{year}{2015}.
\newblock \bibinfo{title}{Is object localization for free? - weakly-supervised
  learning with convolutional neural networks}, in: \bibinfo{booktitle}{2015
  IEEE Conference on Computer Vision and Pattern Recognition (CVPR)}, pp.
  \bibinfo{pages}{685--694}.
\newblock \DOIprefix\doi{10.1109/CVPR.2015.7298668}.
\bibitem[{Pan et~al.(2021)Pan, Gao, Lin, Tang, Dong, Yuan, Huang and
  Xu}]{Pan2021}
\bibinfo{author}{Pan, X.}, \bibinfo{author}{Gao, Y.}, \bibinfo{author}{Lin,
  Z.}, \bibinfo{author}{Tang, F.}, \bibinfo{author}{Dong, W.},
  \bibinfo{author}{Yuan, H.}, \bibinfo{author}{Huang, F.}, \bibinfo{author}{Xu,
  C.}, \bibinfo{year}{2021}.
\newblock \bibinfo{title}{Unveiling the potential of structure preserving for
  weakly supervised object localization}, in: \bibinfo{booktitle}{2021 IEEE/CVF
  Conference on Computer Vision and Pattern Recognition (CVPR)}, pp.
  \bibinfo{pages}{11637--11646}.
\newblock \DOIprefix\doi{10.1109/CVPR46437.2021.01147}.
\bibitem[{Park et~al.(2023)Park, Lee, Lee and Kang}]{fdcnet}
\bibinfo{author}{Park, S.}, \bibinfo{author}{Lee, T.}, \bibinfo{author}{Lee,
  Y.}, \bibinfo{author}{Kang, B.}, \bibinfo{year}{2023}.
\newblock \bibinfo{title}{Fdcnet: Feature drift compensation network for
  class-incremental weakly supervised object localization}, in:
  \bibinfo{booktitle}{Proceedings of the 31st ACM International Conference on
  Multimedia}, \bibinfo{publisher}{Association for Computing Machinery},
  \bibinfo{address}{New York, NY, USA}.
\bibitem[{Russakovsky et~al.(2015)Russakovsky, Deng, Su, Krause, Satheesh, Ma,
  Huang, Karpathy, Khosla, Bernstein, Berg and Fei-Fei}]{ILSVRC15}
\bibinfo{author}{Russakovsky, O.}, \bibinfo{author}{Deng, J.},
  \bibinfo{author}{Su, H.}, \bibinfo{author}{Krause, J.},
  \bibinfo{author}{Satheesh, S.}, \bibinfo{author}{Ma, S.},
  \bibinfo{author}{Huang, Z.}, \bibinfo{author}{Karpathy, A.},
  \bibinfo{author}{Khosla, A.}, \bibinfo{author}{Bernstein, M.},
  \bibinfo{author}{Berg, A.C.}, \bibinfo{author}{Fei-Fei, L.},
  \bibinfo{year}{2015}.
\newblock \bibinfo{title}{{ImageNet Large Scale Visual Recognition Challenge}}.
\newblock \bibinfo{journal}{International Journal of Computer Vision (IJCV)}
  \bibinfo{volume}{115}, \bibinfo{pages}{211--252}.
\newblock \DOIprefix\doi{10.1007/s11263-015-0816-y}.
\bibitem[{Selvaraju et~al.(2017)Selvaraju, Cogswell, Das, Vedantam, Parikh and
  Batra}]{gradcam2017}
\bibinfo{author}{Selvaraju, R.R.}, \bibinfo{author}{Cogswell, M.},
  \bibinfo{author}{Das, A.}, \bibinfo{author}{Vedantam, R.},
  \bibinfo{author}{Parikh, D.}, \bibinfo{author}{Batra, D.},
  \bibinfo{year}{2017}.
\newblock \bibinfo{title}{Grad-cam: Visual explanations from deep networks via
  gradient-based localization}, in: \bibinfo{booktitle}{2017 IEEE International
  Conference on Computer Vision (ICCV)}, pp. \bibinfo{pages}{618--626}.
\newblock \DOIprefix\doi{10.1109/ICCV.2017.74}.
\bibitem[{Singh and Lee(2017)}]{hideseek2017}
\bibinfo{author}{Singh, K.K.}, \bibinfo{author}{Lee, Y.J.},
  \bibinfo{year}{2017}.
\newblock \bibinfo{title}{Hide-and-seek: Forcing a network to be meticulous for
  weakly-supervised object and action localization}, in:
  \bibinfo{booktitle}{2017 IEEE International Conference on Computer Vision
  (ICCV)}, pp. \bibinfo{pages}{3544--3553}.
\newblock \DOIprefix\doi{10.1109/ICCV.2017.381}.
\bibitem[{Szegedy et~al.(2016)Szegedy, Vanhoucke, Ioffe, Shlens and
  Wojna}]{inceptionv3}
\bibinfo{author}{Szegedy, C.}, \bibinfo{author}{Vanhoucke, V.},
  \bibinfo{author}{Ioffe, S.}, \bibinfo{author}{Shlens, J.},
  \bibinfo{author}{Wojna, Z.}, \bibinfo{year}{2016}.
\newblock \bibinfo{title}{Rethinking the inception architecture for computer
  vision}, in: \bibinfo{booktitle}{2016 IEEE Conference on Computer Vision and
  Pattern Recognition (CVPR)}, pp. \bibinfo{pages}{2818--2826}.
\newblock \DOIprefix\doi{10.1109/CVPR.2016.308}.
\bibitem[{Wah et~al.(2011)Wah, Branson, Welinder, Perona and
  Belongie}]{cub_dataset}
\bibinfo{author}{Wah, C.}, \bibinfo{author}{Branson, S.},
  \bibinfo{author}{Welinder, P.}, \bibinfo{author}{Perona, P.},
  \bibinfo{author}{Belongie, S.}, \bibinfo{year}{2011}.
\newblock \bibinfo{title}{The Caltech-UCSD Birds-200-2011 Dataset}.
\newblock \bibinfo{type}{Technical Report} \bibinfo{number}{CNS-TR-2011-001}.
  California Institute of Technology.
\bibitem[{Wei et~al.(2021)Wei, Wang, Li, Wang, Zhou and Cui}]{Wei2021}
\bibinfo{author}{Wei, J.}, \bibinfo{author}{Wang, Q.}, \bibinfo{author}{Li,
  Z.}, \bibinfo{author}{Wang, S.}, \bibinfo{author}{Zhou, S.K.},
  \bibinfo{author}{Cui, S.}, \bibinfo{year}{2021}.
\newblock \bibinfo{title}{Shallow feature matters for weakly supervised object
  localization}, in: \bibinfo{booktitle}{2021 IEEE/CVF Conference on Computer
  Vision and Pattern Recognition (CVPR)}, pp. \bibinfo{pages}{5989--5997}.
\newblock \DOIprefix\doi{10.1109/CVPR46437.2021.00593}.
\bibitem[{Wei et~al.(2019)Wei, Zhang, Wu, Shen and Zhou}]{DDT2019}
\bibinfo{author}{Wei, X.S.}, \bibinfo{author}{Zhang, C.L.},
  \bibinfo{author}{Wu, J.}, \bibinfo{author}{Shen, C.}, \bibinfo{author}{Zhou,
  Z.H.}, \bibinfo{year}{2019}.
\newblock \bibinfo{title}{Unsupervised object discovery and co-localization by
  deep descriptor transformation}.
\newblock \bibinfo{journal}{Pattern Recognition} \bibinfo{volume}{88},
  \bibinfo{pages}{113--126}.
\newblock \DOIprefix\doi{https://doi.org/10.1016/j.patcog.2018.10.022}.
\bibitem[{Woo et~al.(2018)Woo, Park, Lee and Kweon}]{cbam2018}
\bibinfo{author}{Woo, S.}, \bibinfo{author}{Park, J.}, \bibinfo{author}{Lee,
  J.Y.}, \bibinfo{author}{Kweon, I.S.}, \bibinfo{year}{2018}.
\newblock \bibinfo{title}{Cbam: Convolutional block attention module}, in:
  \bibinfo{editor}{Ferrari, V.}, \bibinfo{editor}{Hebert, M.},
  \bibinfo{editor}{Sminchisescu, C.}, \bibinfo{editor}{Weiss, Y.} (Eds.),
  \bibinfo{booktitle}{Computer Vision -- ECCV 2018},
  \bibinfo{publisher}{Springer International Publishing},
  \bibinfo{address}{Cham}. pp. \bibinfo{pages}{3--19}.
\bibitem[{Wu et~al.(2022a)Wu, Zhai and Cao}]{Wu_2022_CVPR}
\bibinfo{author}{Wu, P.}, \bibinfo{author}{Zhai, W.}, \bibinfo{author}{Cao,
  Y.}, \bibinfo{year}{2022}a.
\newblock \bibinfo{title}{Background activation suppression for weakly
  supervised object localization}, in: \bibinfo{booktitle}{Proceedings of the
  IEEE/CVF Conference on Computer Vision and Pattern Recognition (CVPR)}, pp.
  \bibinfo{pages}{14248--14257}.
\bibitem[{Wu et~al.(2022b)Wu, Xu, Wang, Sun, Tan and Weise}]{WU202223_r2}
\bibinfo{author}{Wu, Z.Z.}, \bibinfo{author}{Xu, J.}, \bibinfo{author}{Wang,
  Y.}, \bibinfo{author}{Sun, F.}, \bibinfo{author}{Tan, M.},
  \bibinfo{author}{Weise, T.}, \bibinfo{year}{2022}b.
\newblock \bibinfo{title}{Hierarchical fusion and divergent activation based
  weakly supervised learning for object detection from remote sensing images}.
\newblock \bibinfo{journal}{Information Fusion} \bibinfo{volume}{80},
  \bibinfo{pages}{23--43}.
\newblock \URLprefix
  \url{https://www.sciencedirect.com/science/article/pii/S1566253521002189},
  \DOIprefix\doi{https://doi.org/10.1016/j.inffus.2021.10.010}.
\bibitem[{Xie et~al.(2021)Xie, Luo, Zhu, Jin, Lu and Shen}]{Xie2021}
\bibinfo{author}{Xie, J.}, \bibinfo{author}{Luo, C.}, \bibinfo{author}{Zhu,
  X.}, \bibinfo{author}{Jin, Z.}, \bibinfo{author}{Lu, W.},
  \bibinfo{author}{Shen, L.}, \bibinfo{year}{2021}.
\newblock \bibinfo{title}{Online refinement of low-level feature based
  activation map for weakly supervised object localization}, in:
  \bibinfo{booktitle}{2021 IEEE/CVF International Conference on Computer Vision
  (ICCV)}, pp. \bibinfo{pages}{132--141}.
\newblock \DOIprefix\doi{10.1109/ICCV48922.2021.00020}.
\bibitem[{Xie et~al.(2022)Xie, Xiang, Chen, Hou, Zhao and Shen}]{Xie_2022_CVPR}
\bibinfo{author}{Xie, J.}, \bibinfo{author}{Xiang, J.}, \bibinfo{author}{Chen,
  J.}, \bibinfo{author}{Hou, X.}, \bibinfo{author}{Zhao, X.},
  \bibinfo{author}{Shen, L.}, \bibinfo{year}{2022}.
\newblock \bibinfo{title}{C2am: Contrastive learning of class-agnostic
  activation map for weakly supervised object localization and semantic
  segmentation}, in: \bibinfo{booktitle}{Proceedings of the IEEE/CVF Conference
  on Computer Vision and Pattern Recognition (CVPR)}, pp.
  \bibinfo{pages}{989--998}.
\bibitem[{Xue et~al.(2019)Xue, Liu, Wan, Jiao, Ji and Ye}]{Xue2019}
\bibinfo{author}{Xue, H.}, \bibinfo{author}{Liu, C.}, \bibinfo{author}{Wan,
  F.}, \bibinfo{author}{Jiao, J.}, \bibinfo{author}{Ji, X.},
  \bibinfo{author}{Ye, Q.}, \bibinfo{year}{2019}.
\newblock \bibinfo{title}{Danet: Divergent activation for weakly supervised
  object localization}, in: \bibinfo{booktitle}{2019 IEEE/CVF International
  Conference on Computer Vision (ICCV)}, pp. \bibinfo{pages}{6588--6597}.
\newblock \DOIprefix\doi{10.1109/ICCV.2019.00669}.
\bibitem[{Yang et~al.(2020)Yang, Kim, Kim and Kim}]{Yang2020}
\bibinfo{author}{Yang, S.}, \bibinfo{author}{Kim, Y.}, \bibinfo{author}{Kim,
  Y.}, \bibinfo{author}{Kim, C.}, \bibinfo{year}{2020}.
\newblock \bibinfo{title}{Combinational class activation maps for weakly
  supervised object localization}, in: \bibinfo{booktitle}{2020 IEEE Winter
  Conference on Applications of Computer Vision (WACV)}, pp.
  \bibinfo{pages}{2930--2938}.
\newblock \DOIprefix\doi{10.1109/WACV45572.2020.9093566}.
\bibitem[{Zhang et~al.(2020a)Zhang, Cao and Wu}]{Zhang2020}
\bibinfo{author}{Zhang, C.L.}, \bibinfo{author}{Cao, Y.H.},
  \bibinfo{author}{Wu, J.}, \bibinfo{year}{2020}a.
\newblock \bibinfo{title}{Rethinking the route towards weakly supervised object
  localization}, in: \bibinfo{booktitle}{2020 IEEE/CVF Conference on Computer
  Vision and Pattern Recognition (CVPR)}, pp. \bibinfo{pages}{13457--13466}.
\newblock \DOIprefix\doi{10.1109/CVPR42600.2020.01347}.
\bibitem[{Zhang et~al.(2022)Zhang, Han, Cheng and Yang}]{Zhang2022_r2}
\bibinfo{author}{Zhang, D.}, \bibinfo{author}{Han, J.}, \bibinfo{author}{Cheng,
  G.}, \bibinfo{author}{Yang, M.H.}, \bibinfo{year}{2022}.
\newblock \bibinfo{title}{Weakly supervised object localization and detection:
  A survey}.
\newblock \bibinfo{journal}{IEEE Transactions on Pattern Analysis and Machine
  Intelligence} \bibinfo{volume}{44}, \bibinfo{pages}{5866--5885}.
\newblock \DOIprefix\doi{10.1109/TPAMI.2021.3074313}.
\bibitem[{Zhang et~al.(2018a)Zhang, Wei, Feng, Yang and Huang}]{ACoL2018}
\bibinfo{author}{Zhang, X.}, \bibinfo{author}{Wei, Y.}, \bibinfo{author}{Feng,
  J.}, \bibinfo{author}{Yang, Y.}, \bibinfo{author}{Huang, T.},
  \bibinfo{year}{2018}a.
\newblock \bibinfo{title}{Adversarial complementary learning for weakly
  supervised object localization}, in: \bibinfo{booktitle}{2018 IEEE/CVF
  Conference on Computer Vision and Pattern Recognition}, pp.
  \bibinfo{pages}{1325--1334}.
\newblock \DOIprefix\doi{10.1109/CVPR.2018.00144}.
\bibitem[{Zhang et~al.(2018b)Zhang, Wei, Kang, Yang and Huang}]{spg2018}
\bibinfo{author}{Zhang, X.}, \bibinfo{author}{Wei, Y.}, \bibinfo{author}{Kang,
  G.}, \bibinfo{author}{Yang, Y.}, \bibinfo{author}{Huang, T.},
  \bibinfo{year}{2018}b.
\newblock \bibinfo{title}{Self-produced guidance for weakly-supervised object
  localization}, in: \bibinfo{editor}{Ferrari, V.}, \bibinfo{editor}{Hebert,
  M.}, \bibinfo{editor}{Sminchisescu, C.}, \bibinfo{editor}{Weiss, Y.} (Eds.),
  \bibinfo{booktitle}{Computer Vision -- ECCV 2018},
  \bibinfo{publisher}{Springer International Publishing},
  \bibinfo{address}{Cham}. pp. \bibinfo{pages}{610--625}.
\bibitem[{Zhang et~al.(2020b)Zhang, Wei and Yang}]{i2c2020}
\bibinfo{author}{Zhang, X.}, \bibinfo{author}{Wei, Y.}, \bibinfo{author}{Yang,
  Y.}, \bibinfo{year}{2020}b.
\newblock \bibinfo{title}{Inter-image communication for weakly supervised
  localization}, in: \bibinfo{editor}{Vedaldi, A.}, \bibinfo{editor}{Bischof,
  H.}, \bibinfo{editor}{Brox, T.}, \bibinfo{editor}{Frahm, J.M.} (Eds.),
  \bibinfo{booktitle}{Computer Vision -- ECCV 2020},
  \bibinfo{publisher}{Springer International Publishing},
  \bibinfo{address}{Cham}. pp. \bibinfo{pages}{271--287}.
\bibitem[{Zhou et~al.(2016)Zhou, Khosla, Lapedriza, Oliva and
  Torralba}]{cam2016}
\bibinfo{author}{Zhou, B.}, \bibinfo{author}{Khosla, A.},
  \bibinfo{author}{Lapedriza, A.}, \bibinfo{author}{Oliva, A.},
  \bibinfo{author}{Torralba, A.}, \bibinfo{year}{2016}.
\newblock \bibinfo{title}{Learning deep features for discriminative
  localization}, in: \bibinfo{booktitle}{2016 IEEE Conference on Computer
  Vision and Pattern Recognition (CVPR)}, pp. \bibinfo{pages}{2921--2929}.
\newblock \DOIprefix\doi{10.1109/CVPR.2016.319}.
\bibitem[{Zhu et~al.(2022)Zhu, She, Chen, You, Wang and Lu}]{Zhu_2022_CVPR}
\bibinfo{author}{Zhu, L.}, \bibinfo{author}{She, Q.}, \bibinfo{author}{Chen,
  Q.}, \bibinfo{author}{You, Y.}, \bibinfo{author}{Wang, B.},
  \bibinfo{author}{Lu, Y.}, \bibinfo{year}{2022}.
\newblock \bibinfo{title}{Weakly supervised object localization as domain
  adaption}, in: \bibinfo{booktitle}{Proceedings of the IEEE/CVF Conference on
  Computer Vision and Pattern Recognition (CVPR)}, pp.
  \bibinfo{pages}{14637--14646}.

\end{thebibliography}

\end{document}